\documentclass[10pt,twocolumn,letterpaper]{article}
\ifdefined\pdfminorversion\pdfminorversion=7\fi
\usepackage[pagenumbers]{cvpr}
\usepackage[T1]{fontenc}
\usepackage{colortbl}

\definecolor{rgGainGreen}{RGB}{0,113,75}
\newcommand{\rgGain}[2]{\textbf{#1}\,\raisebox{.45ex}{\scriptsize\textcolor{rgGainGreen}{$\blacktriangle$#2}}}

\usepackage{microtype}
\usepackage{cuted}
\usepackage{float}
\definecolor{cvprblue}{rgb}{0.21,0.49,0.74}
\usepackage[breaklinks,colorlinks,allcolors=cvprblue]{hyperref}
\def\confName{CVPR}
\def\confYear{2027}
\title{RelGeo: Pair-Conditioned Relational Calibration for Frozen Diffusion Correspondence}
\author{%
Zixuan Dai$^{1,2}$ \quad Zhiliang Zhang$^{1,2}$ \quad Daosong Hu$^{1}$ \quad Tang Zheng$^{1,2}$\\[3pt]
Xinkai Liu$^{1,2}$ \quad Tianyu Lei$^{1,2}$ \quad Wei Zhan$^{1,2,*}$\\[5pt]
{\normalsize $^{1}$School of Computer Science, Yangtze University}\\[2pt]
{\normalsize $^{2}$Yangtze University Interdisciplinary Innovation Research Center for Smart Agriculture, Yangtze University}
}
\hypersetup{pdftitle={RelGeo: Pair-Conditioned Relational Calibration for Frozen Diffusion Correspondence},pdfauthor={Zixuan Dai, Zhiliang Zhang, Daosong Hu, Tang Zheng, Xinkai Liu, Tianyu Lei, Wei Zhan}}
\begin{document}
\maketitle
\vspace{-22pt} 
\begingroup
\renewcommand{\thefootnote}{}
\footnotetext{\textsuperscript{*} Corresponding author.}
\endgroup
\begin{strip}
\centering
\includegraphics[width=0.7\textwidth]{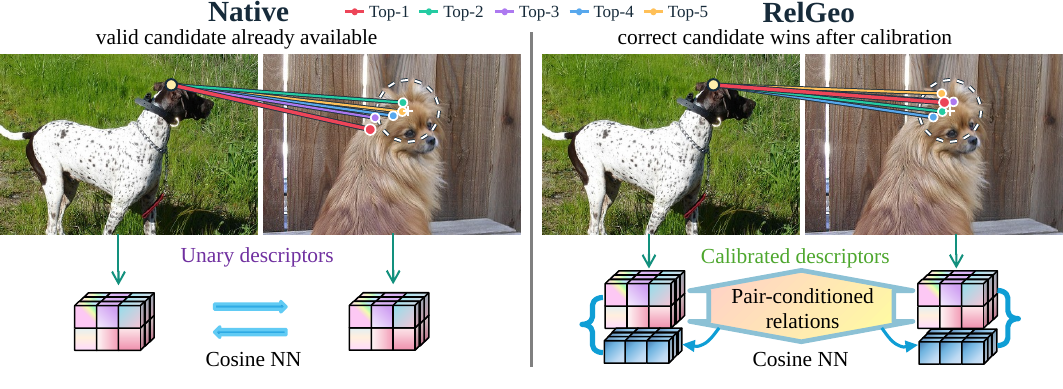}
\captionsetup{hypcap=false}
\captionof{figure}{Native cosine matching selects an incorrect location despite a valid alternative already present among its Top-5 peaks. RelGeo injects pair-conditioned relations into descriptor matching, correcting candidate selection while keeping cosine nearest-neighbor readout.}
\label{fig:teaser}
\captionsetup{hypcap=true}
\end{strip}
\par

\begin{abstract}
Frozen diffusion descriptors retrieve strong semantic correspondence candidates, yet cosine nearest-neighbor matching can select incorrect locations despite valid high-ranking alternatives. Across DIFT-SD2.1 and DiTF-FLUX on SPair-71k, AP-10K, and SOCO (CC, SOC, Cross-SOC), 41--69\% of Top-1 errors retain a PCK@0.10-valid match among the original Top-5 spatial peaks, revealing a persistent candidate-selection bottleneck. We find that descriptor rankings and current-pair relational evidence can favor different candidates: reciprocal diffusion Query/Key associations and pretrained RoMa transport exhibit distinct, partly complementary preferences. To incorporate this evidence into matching decisions, we introduce RelGeo, a Pair-Conditioned Relational Calibration framework expressing heterogeneous relations as shared source--target coordinates. Spectral Relational Coordinates (SRC) encode reciprocal diffusion affinities, while Warp Identity Coordinates (WIC) encode reliability-weighted bidirectional transport endpoints. Their inner products augment frozen descriptor similarity while preserving full-domain cosine nearest-neighbor matching without additional semantic-correspondence training. RelGeo improves all ten evaluations by an average of 5.51 PCK@0.10 points; fixed-candidate reranking and pair-misalignment controls corroborate improved selection and dependence on current-pair relational alignment.
\end{abstract}

\section{Introduction}
\label{sec:introduction}

Cross-image semantic correspondence establishes semantically consistent matches across changes in appearance, viewpoint, and pose~\cite{ref18,ref45}, supporting image editing~\cite{ref5}, video object segmentation~\cite{refVideoMatch2018}, and object tracking~\cite{refOmniMotion2023}. Dense features from large-scale pretrained vision models~\cite{ref26,ref27} have enabled strong matching performance without correspondence-specific training~\cite{ref1,ref5}. Since DIFT~\cite{ref1}, diffusion correspondence has advanced through improved feature extraction~\cite{ref3,ref25}, aggregation~\cite{ref2,ref56}, fusion~\cite{ref5,ref48}, and increasingly discriminative Diffusion Transformer representations in DiTF~\cite{ref4,ref44,ref57,ref58}. Despite these advances, frozen diffusion correspondence still commonly relies on cosine nearest-neighbor matching. A correct prediction requires both retrieving a valid candidate and ranking it above competing locations. This raises a question worth exploring: as diffusion descriptors become stronger, do their performance gains stem more from retrieving valid candidates or from selecting more reliably among those already retrieved?

\begin{figure}[!t]
\centering
\includegraphics[width=0.94\linewidth,trim={0 9bp 0 6bp},clip]{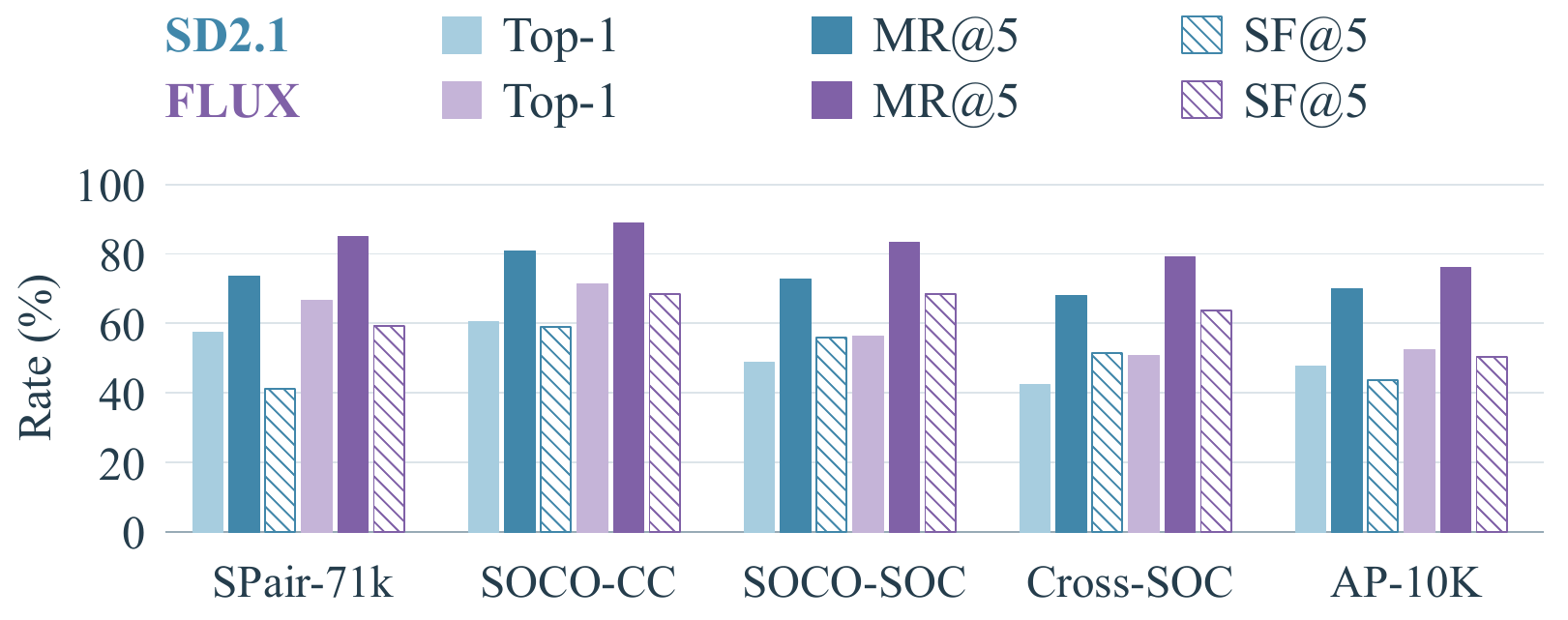}
\caption{Candidate availability in Native diffusion similarity responses across five correspondence settings. Top-1 is PCK@0.10; MR@5 measures coverage by at least one valid Top-5 peak; SF@5 measures this coverage conditioned on an incorrect Native Top-1. Candidate peaks are extracted from the original responses.}
\label{fig:candidate_availability}
\end{figure}

To answer this question, we examine the original cosine responses of DIFT-SD2.1 and DiTF-FLUX across five correspondence settings. We find that 41--69\% of incorrect Top-1 predictions still contain a PCK@0.10-valid match among the original Top-5 spatial peaks (Fig.~\ref{fig:candidate_availability}). Taking SOCO-SOC as an example, FLUX increases Top-5 coverage from 72.91\% to 83.45\%, while Top-1 accuracy rises from 48.89\% to 56.58\%, widening their gap from 24.02 to 26.87 percentage points. Figure~\ref{fig:teaser} illustrates a candidate-selection failure: a valid match appears among the Top-5 peaks, but an incorrect location receives the highest cosine score. These findings reveal a persistent \emph{candidate-selection bottleneck} despite improved candidate retrieval.

Why do valid candidates still lose the final match despite being retrieved? We argue that semantic similarity alone cannot fully determine correspondence identity in the current image pair. Diffusion descriptors encode rich semantic and spatial context, but pointwise cosine matching compares features extracted independently from the two images without explicitly testing whether a candidate agrees with their pair-specific correspondence relations. This raises a further question: can relations computed for the current image pair resolve candidate ambiguities left by descriptor similarity? To test this hypothesis, we compare reciprocal diffusion Query/Key associations and geometric transport from pretrained RoMa~\cite{ref9} on the same valid and incorrect candidates identified by Native. Both relational sources favor valid candidates rejected by the original cosine ranking in some cases and exhibit distinct, partly complementary preferences (Fig.~\ref{fig:raw_preferences}; Sec.~\ref{sec:raw_evidence}).

Motivated by these observations, we introduce \textbf{RelGeo} (\emph{Pair-Conditioned Relational Calibration}) to incorporate these relations into frozen descriptor matching. \emph{Spectral Relational Coordinates} (SRC) transform reciprocal diffusion Query/Key affinities into aligned source--target coordinates whose inner products encode pairwise associations. \emph{Warp Identity Coordinates} (WIC) express RoMa's bidirectional transport as reliability-weighted correspondence evidence, using confidence and cycle consistency to attenuate uncertain hypotheses. Both coordinate sources augment semantic descriptors and preserve full-domain cosine nearest-neighbor matching without additional semantic-correspondence training or per-pair optimization (Fig.~\ref{fig:method}).

Our main contributions are:
\begin{itemize}
    \setlength{\itemsep}{0pt}
    \setlength{\parsep}{0pt}
    \setlength{\topsep}{3pt}
    \setlength{\partopsep}{0pt}
    \item We distinguish candidate retrieval from final selection, revealing a persistent gap between valid-candidate availability and Top-1 accuracy across two diffusion pipelines and five settings.
    \item We identify distinct, partly complementary relational evidence in reciprocal diffusion associations and pretrained geometric transport by comparing identical competing candidates.
    \item We propose RelGeo to calibrate frozen descriptor similarity with pair-conditioned relations while preserving full-domain cosine matching. It improves all ten evaluations by an average of \textbf{5.51 PCK@0.10 points}; fixed-candidate reranking and pair-misalignment controls support improved selection and current-pair dependence.
\end{itemize}

\section{Related Work}
\label{sec:related_work}

\vspace{2pt}\noindent\textbf{Diffusion Descriptors and Feature Fusion.}\enspace
Semantic correspondence has progressed from task-specific features and matching-cost aggregation~\cite{ref18,ref23,ref15,ref16} to transferable pretrained representations~\cite{ref26,ref27,ref41}. Diffusion correspondence uses pretrained features extracted by DIFT~\cite{ref1} and query-conditioned attention through prompt optimization~\cite{ref53}. Subsequent work aggregates features across layers and timesteps~\cite{ref2}, models their temporal evolution as diffusion trajectories~\cite{ref59}, and improves extraction through noise removal and distillation~\cite{ref3,ref25}. DiTF and related analyses enhance spatial discrimination by examining and modulating massive activations in diffusion Transformers~\cite{ref4,refDG2026,ref52}; structure-aware aggregation preserves coarse structure while selectively aggregating fine details~\cite{ref56}. Another line combines complementary pretrained features: A Tale of Two Features fuses diffusion and DINO representations~\cite{ref5}, MATCHA learns semantic--geometric feature integration~\cite{ref20}, Fuse2Match combines frozen flow, diffusion, and contrastive features~\cite{ref48}, and SimBetter protects feature structure while modeling feature interactions~\cite{ref60}. These methods strengthen correspondence representations, while improvements in final accuracy alone do not separate gains in candidate retrieval from gains in candidate selection.

\vspace{2pt}\noindent\textbf{Geometric Structure and Identity.}\enspace
Symmetry, repeated parts, and viewpoint changes can leave multiple semantically plausible correspondences with different identities. GeoAware~\cite{ref6} identifies this geometric ambiguity in foundation-model features and addresses it through viewpoint alignment and geometry-aware feature refinement. Viewpoint-guided representations~\cite{ref10} and 3D geometric reasoning and alignment~\cite{ref11,ref12} provide further structural cues. GECO~\cite{ref61} learns geometrically coherent embeddings through optimal-transport-based supervision, while Shape-of-You~\cite{ref7} combines feature similarity and 3D structural consistency through fused Gromov--Wasserstein transport to generate correspondence pseudo-labels. SOCO~\cite{ref8} distinguishes local semantic concepts, object-relative identities, and cross-category structural correspondence, enabling more targeted evaluation of identity errors. These approaches establish the value of geometry and structure; RelGeo examines the complementary question of how to choose the correct identity when frozen diffusion descriptors have already retrieved a valid candidate.

\vspace{2pt}\noindent\textbf{Pair-Conditioned and Structured Matching.}\enspace
Pairwise compatibility has been modeled through neighborhood consensus~\cite{ref22,ref43}, geometric voting~\cite{ref14}, cost aggregation~\cite{ref15,ref16}, and global match-to-match attention~\cite{refTransforMatcher2022}. Spectral matching~\cite{ref38} and optimal transport~\cite{ref13,ref51} impose structured matching constraints. Functional maps, originating in shape correspondence~\cite{ref50}, have also been used to optimize globally coherent mappings of pretrained image features~\cite{ref49}. Learned cross-image interaction~\cite{ref32,refACTR2023} and pair-conditioned diffusion representations~\cite{ref17,ref21} further incorporate information specific to an image pair. Dense matchers such as DKM~\cite{ref33} and RoMa~\cite{ref9} predict geometric transport. Closely related to the candidate-selection perspective, Preserve, Then Resolve~\cite{ref62} retains high-ranking associations from frozen visual features and resolves them through robust estimation for relative pose recovery. RelGeo instead uses reciprocal frozen diffusion Query/Key associations and pretrained bidirectional transport as aligned source--target coordinates, calibrating full-domain cosine matching without separate global map optimization or a robust-estimation stage.

\section{From Candidate Formation to Correspondence Decisions}
\label{sec:selection_failures}

\subsection{Candidate Availability versus Selection}

We denote the original cosine response by \emph{Native} and extract its five highest spatial peaks using non-maximum suppression (NMS). Let $E$ denote an incorrect Native Top-1 prediction and $A_K$ indicate that at least one of the first $K$ peaks satisfies PCK@0.10. MR@$K=\Pr(A_K)$ measures valid-candidate coverage and gives the oracle upper bound for reranking these peaks. SF@5 $=\Pr(A_5\mid E)$ measures the fraction of Native errors with an available valid candidate. Since Top-1 belongs to the candidate set, MR@5 minus Top-1 PCK quantifies candidate-available selection errors under the same benchmark aggregation (Supplementary S1 and S3).

This distinction partitions Native predictions into correct selections, candidate-available errors, and candidate-missing errors. On SOCO-SOC, moving from SD2.1 to FLUX reduces the candidate-missing error rate from 27.09\% to 16.55\%, while candidate-available selection errors increase from 24.02\% to 26.87\%. Thus, the 10.54-point reduction in missing-candidate errors is accompanied by a 2.85-point increase in errors where a valid candidate remains unselected, yielding a net Top-1 gain of 7.69 points. The aggregate Top-1/MR@5 ratio likewise changes only from 67.06\% to 67.80\%. Stronger descriptors substantially improve candidate retrieval, whereas the selection of available candidates advances much less. This decomposition identifies candidate selection as a persistent source of correspondence errors beyond descriptor quality.

\subsection{Same-Concept Competition and Correspondence Identity}

Repeated parts and viewpoint changes can produce multiple plausible target matches with different correspondence identities. SOCO~\cite{ref8} distinguishes semantic concepts from object-relative identity, enabling an analysis of competition among semantically similar locations.

Let $M_{\mathrm{ann}}$ count visible target annotations sharing the source concept, including the designated counterpart. Across the 97 SOCO-SOC classes represented in both $M_{\mathrm{ann}}=1$ and $M_{\mathrm{ann}}\geq2$, the class-macro joint rate $P(E\cap A_5)$ increases from 18.24\% to 26.33\% for SD2.1 and from 19.29\% to 31.01\% for FLUX. Meanwhile, MR@5 decreases by 4.50 and 6.18 points, respectively (Supplementary S3). Greater same-concept competition thus accompanies more selection errors despite the availability of valid candidates.

\begin{figure}[!t]
\centering
\includegraphics[width=0.9\linewidth,trim={0 3bp 0 2bp},clip]{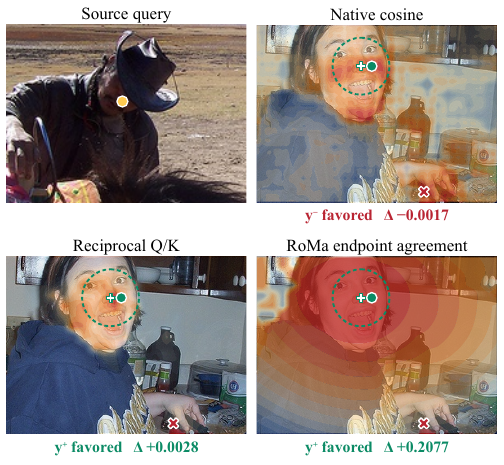}
\caption{Raw preferences for a face query on SPair-71k. Native favors the hand-adjacent peak $y^-$. Both relations favor the valid face peak $y^+$ (Native rank 2). $\Delta=s(y^+)-s(y^-)$ uses each signal's own scale. The ring marks the PCK region.}
\label{fig:raw_preferences}
\end{figure}

Correspondence identity also reveals a distinction between overall matching accuracy and the discrimination of competing locations. Although FLUX achieves higher full-domain PCK, its accuracy in ranking annotated same-concept locations is lower than SD2.1 (59.66\% versus 63.17\%; Supplementary S7). The latter evaluates selection among annotated alternatives, while full-domain PCK reflects the maximum response across the target image. A valid counterpart may therefore outrank other same-concept annotations yet lose to an unrelated high-response location. Figure~\ref{fig:raw_preferences} illustrates this broader ambiguity: a face query is attracted to a hand-adjacent distractor. Candidate-selection errors extend beyond symmetric or identically labeled parts.

\begin{figure*}[!t]
\centering
\includegraphics[width=0.9\textwidth,trim={0 0 0 10bp},clip]{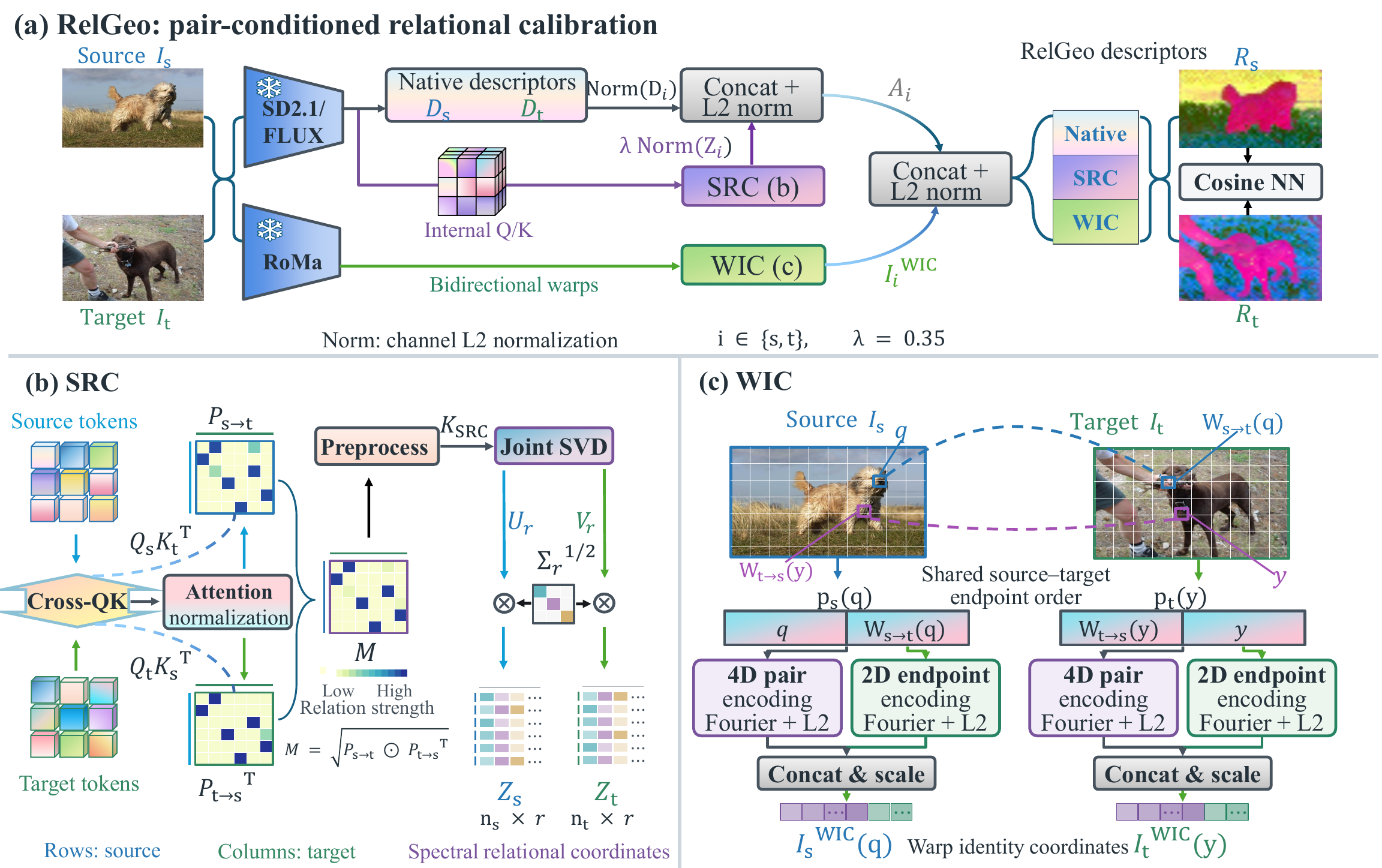}
\caption{Overview of RelGeo. Frozen diffusion descriptors are augmented with two pair-conditioned relational coordinate maps for full-domain cosine matching. SRC constructs shared spectral coordinates from reciprocal Q/K associations, while WIC encodes bidirectional transport hypotheses in a common source--target endpoint order, weighted by transport confidence and cycle consistency.}
\label{fig:method}
\end{figure*}

\subsection{Relational Evidence for Competing Candidates}
\label{sec:raw_evidence}
\label{sec:reference_hypothesis}

To examine whether current-pair relations provide additional evidence for candidate selection, we fix Native's incorrect maximum $y^-$ and the highest-ranked valid Top-5 peak $y^+$ for each eligible error. All signals are compared on these same two locations, separating candidate discrimination from changes in retrieval.

We examine two relational sources before constructing RelGeo coordinates. The internal signal measures reciprocal cross-image Query/Key association reconstructed from frozen diffusion states. The external signal measures agreement between RoMa's forward projection of the source query and backward projection of the target candidate. Higher reciprocal association or smaller summed endpoint residual favors a candidate. These measurements precede SRC spectral factorization and WIC encoding (Supplementary S4).

On SOCO-SOC, raw Q/K associations favor the valid peak in 50.64\%/39.80\% of eligible SD2.1/FLUX comparisons, compared with 72.84\%/67.68\% for geometric transport. On AP-10K, the corresponding rates are 43.64\%/40.53\% and 59.18\%/56.92\%. The two sources also disagree on substantial subsets. On SOCO-SOC, Q/K favors the valid candidate while transport favors the incorrect one in 11.74\%/10.14\% of comparisons; the reverse occurs in 33.94\%/38.02\% (Supplementary S4). Figure~\ref{fig:raw_preferences} illustrates a case where both relations favor the valid face peak despite Native's incorrect ranking.

These comparisons reveal distinct and partly complementary relational evidence. Geometric transport more frequently favors valid candidates, while internal diffusion associations retain useful discrimination when transport favors an incorrect location. Their combined evidence motivates pair-conditioned calibration of the original semantic similarity, allowing candidate ranking to reflect both descriptor agreement and relations specific to the current image pair.

\section{Pair-Conditioned Relational Calibration}
\label{sec:method}

\subsection{Overview}

Section~\ref{sec:selection_failures} identifies selection errors despite valid retrieved candidates and shows that current-pair relations can distinguish some competing locations. RelGeo incorporates these relations into the original descriptor comparison, allowing semantic similarity and pair-specific evidence to determine the final match.

Given source and target images $I_s$ and $I_t$, a frozen diffusion model independently extracts descriptor maps $D_s$ and $D_t$. For source query $q$ and target domain $\Omega_t$, Native predicts
\begin{equation}
\hat y_{\mathrm{Native}}(q)
=\operatorname*{arg\,max}_{y\in\Omega_t}
\left\langle
\mathcal N(D_s(q)),
\mathcal N(D_t(y))
\right\rangle,
\label{eq:1}
\end{equation}
where $\mathcal N$ denotes per-location $\ell_2$ normalization.

RelGeo augments the descriptors with two aligned source--target coordinate maps (Fig.~\ref{fig:method}): Spectral Relational Coordinates (SRC) encode reciprocal diffusion Query/Key associations, and Warp Identity Coordinates (WIC) encode reliability-weighted geometric transport. Both contribute to the same cosine readout. The Top-5 candidates in Sec.~\ref{sec:selection_failures} are used for analysis; RelGeo matches over the full target domain.

\subsection{Internal Spectral Relational Coordinates}

\emph{Recovering relational information from frozen diffusion models.}
Diffusion descriptors encode semantic and spatial context, but their pointwise cosine comparison does not explicitly use cross-image associations recoverable from frozen Query/Key projections. Comparing source and target projections provides a complementary measure of candidate compatibility, as observed in Sec.~\ref{sec:raw_evidence}.

We reconstruct cross-image affinities from independently extracted Query/Key states using $Q_sK_t^\top$ and $Q_tK_s^\top$. Scaled logits are normalized while retaining the original-context attention mass, then averaged across attention heads and sampling members, yielding $P_{s\to t}$ and $P_{t\to s}$. We combine the two directions as
\begin{equation}
M=\sqrt{P_{s\to t}\odot P_{t\to s}^{\top}},
\qquad
K^{\mathrm{SRC}}=\Psi_{\pi}(M),
\label{eq:2}
\end{equation}
where $\odot$ and the square root are element-wise. Reciprocal agreement emphasizes associations supported in both directions. The fixed preprocessing $\Psi_{\pi}$ (configuration $\pi$ varies by backbone and benchmark) adjusts row/column affinity backgrounds and, for FLUX, incorporates local association support (Supplementary S2).

\emph{Why spectral coordinates?}
The resulting affinity kernel describes relations between pairs of locations, whereas cosine matching operates on a descriptor at each location. SRC connects these two representations through the rank-$r$ truncated SVD of $K^{\mathrm{SRC}}$: $K_r$ is its approximation, $U_r$, $V_r$ its singular-vector matrices, and $\Sigma_r$ its diagonal singular values.
\begin{equation}
\begin{aligned}
K_r &= U_r\Sigma_rV_r^\top,\\
Z_s &= U_r\Sigma_r^{1/2},
\qquad
Z_t = V_r\Sigma_r^{1/2}.
\end{aligned}
\label{eq:3}
\end{equation}

The joint factorization yields paired source--target coordinates satisfying $Z_sZ_t^\top=K_r$ before normalization and resampling. Their dot products recover the dominant pairwise associations, rather than introducing unrelated feature dimensions. Because both maps arise from the same factorization, corresponding components remain aligned across images. We resample and normalize $Z_s$ and $Z_t$ for fusion.

\subsection{External Warp Identity Coordinates}

\emph{Why use pretrained geometric transport?}
Internal Query/Key associations provide useful evidence, but do not consistently favor the valid candidate (Sec.~\ref{sec:raw_evidence}). We therefore use an external relation source trained on paired images. RoMa~\cite{ref9} is pretrained for dense geometric matching with correspondence supervision on MegaDepth. Its frozen forward and backward transport fields provide pair-specific hypotheses that complement diffusion associations.

\emph{From geometric predictions to semantic correspondence evidence.}
Geometric matching and cross-instance semantic correspondence need not share the same endpoint: changes in pose, instance, and visible structure can make a plausible geometric prediction semantically incorrect. Direct RoMa prediction selects its estimated endpoint, while a Gaussian prior around the forward endpoint favors nearby locations without evaluating the reverse mapping. WIC instead treats these predictions as \emph{correspondence hypotheses} and scores candidate compatibility in both directions. Transport reliability controls the influence of this evidence on semantic matching. Direct RoMa and forward Gaussian fusion are compared in Supplementary S6.

Let $W_{s\to t}$ and $W_{t\to s}$ denote the frozen forward and backward transport fields. Using normalized image-grid coordinates $g_s(q)$ and $g_t(y)$, we represent a source query $q$ and target candidate $y$ in a common source--target order:
\begin{equation}
\begin{aligned}
p_s(q)&=[g_s(q),W_{s\to t}(q)],\\
p_t(y)&=[W_{t\to s}(y),g_t(y)].
\end{aligned}
\label{eq:4}
\end{equation}

A consistent match satisfies $W_{s\to t}(q)\approx g_t(y)$ and $W_{t\to s}(y)\approx g_s(q)$.

WIC turns this agreement into comparable coordinates using a fixed, unit-normalized Fourier encoding $\phi$~\cite{ref39}:
\begin{equation}
\begin{aligned}
I_s^{\mathrm{WIC}}(q)
&=\frac{\rho_s(q)}{\sqrt{2}}
\left[
\phi(p_s(q)),
\phi(W_{s\to t}(q))
\right],\\
I_t^{\mathrm{WIC}}(y)
&=\frac{\rho_t(y)}{\sqrt{2}}
\left[
\phi(p_t(y)),
\phi(g_t(y))
\right].
\end{aligned}
\label{eq:wic}
\end{equation}

The first block scores agreement between the complete bidirectional endpoint hypotheses, providing a soft correspondence-identity cue. The second preserves target-domain proximity when backward transport is imperfect. Their inner products assign compatibility to all target candidates without fixing RoMa's forward endpoint.

\emph{Reliability-aware calibration.}
RoMa confidence and forward--backward cycle consistency regulate the contribution of each hypothesis. With normalized confidence support $S_i$ and cycle residual $e_i$ measured in grid-cell units, the reliability amplitude is
\begin{equation}
\rho_i=\frac{\sqrt{S_i}}{1+e_i},
\qquad i\in\{s,t\}.
\label{eq:reliability}
\end{equation}

Since both Fourier blocks are unit-normalized, $\|I_i^{\mathrm{WIC}}\|_2=\rho_i$. Applying reliability after positional encoding preserves its amplitude in the WIC inner product and final cosine normalization. Supplementary S2 specifies confidence normalization, cycle computation, and coordinate conventions.

\subsection{Relational Fusion and Correspondence Readout}

\emph{One matching decision, three sources of evidence.}
SRC and WIC complement the original descriptors rather than replacing their semantic evidence. We combine all three through normalized concatenation, retaining cosine nearest-neighbor matching:
\begin{equation}
\begin{aligned}
A_i
&=\mathcal N\left(
[\mathcal N(D_i),\lambda\mathcal N(Z_i)]
\right),\\
R_i
&=\mathcal N\left(
[\mathcal N(A_i),I_i^{\mathrm{WIC}}]
\right),
\end{aligned}
\qquad i\in\{s,t\}.
\label{eq:5}
\end{equation}
We fix the SRC weight to $\lambda=0.35$.

To make their contributions explicit, let $s_D$ and $s_Z$ denote semantic and SRC cosine similarities, and let $s_W(q,y)=\langle I_s^{\mathrm{WIC}}(q),I_t^{\mathrm{WIC}}(y)\rangle$ be WIC agreement. On the construction grids, before readout interpolation and for nonzero normalized branches,
\begin{equation}
s_R(q,y)=
\frac{
\displaystyle
\frac{s_D(q,y)+\lambda^2s_Z(q,y)}{1+\lambda^2}
+s_W(q,y)
}{
\sqrt{
(1+\rho_s(q)^2)
(1+\rho_t(y)^2)
}
}.
\label{eq:relgeo_score}
\end{equation}

The score combines original similarity, SRC associations, and reliability-weighted transport agreement. A lower-ranked Native candidate can therefore become the final match when the current-pair relations provide stronger support.

The prediction remains the full-domain cosine nearest neighbor:
\begin{equation}
\hat y_{\mathrm{RelGeo}}(q)
=\operatorname*{arg\,max}_{y\in\Omega_t}
\left\langle R_s(q),R_t(y)\right\rangle.
\label{eq:6}
\end{equation}

The procedure requires no additional semantic-correspondence training, pair-specific optimization, or separate candidate-assignment solver.

\section{Experiments}
\label{sec:experiments}

\subsection{Experimental Setup}
\label{sec:eval_settings}

\vspace{2pt}\noindent\textbf{Datasets and evaluation.}\enspace
We evaluate RelGeo on SPair-71k~\cite{ref45}, SOCO~\cite{ref8}, and AP-10K~\cite{ref19}, covering five settings: SPair-71k, SOCO-CC, SOCO-SOC, Cross-SOC, and AP-10K intraspecies matching. SPair-71k spans 18 categories; SOCO distinguishes concept, object-relative identity, and cross-category structure; AP-10K covers animal keypoints. We use percentage of correct keypoints (PCK@0.10) with benchmark-specific normalization. SPair reports category-macro mean per-point PCK and SOCO uses query--pair--class aggregation; sample support and coordinate adapters are detailed in Supplementary S1.

\vspace{2pt}\noindent\textbf{Implementation details.}\enspace
We use frozen DIFT-SD2.1 and DiTF-FLUX pipelines; the latter uses FLUX.1-dev~\cite{ref58}. Within each backbone, Native, single-source variants, and RelGeo share the same images, extraction settings, queries, and cosine readout. $\dagger$ indicates independent DINOv2-B/14 appearance fusion; RoMa~\cite{ref9} itself contains DINOv2-L/14. Supplementary S1--S2 detail pretrained resources and evaluation protocols; S8.2 reports timing with cached RoMa fields.

\subsection{Main Results and Comparison with State-of-the-Art Methods}

\begin{table}[!htbp]
\centering
\caption{Controlled PCK@0.10 (\%) for two backbones across five settings. Green triangles show gains over Native in percentage points. The mean gain across all ten combinations is 5.51 points.}
\label{tab:1}
\small
\renewcommand{\arraystretch}{0.95}
\begin{tabular}{@{}lrrrr@{}}
\toprule
 & \multicolumn{2}{c}{DIFT-SD2.1~\cite{ref1}} & \multicolumn{2}{c}{DiTF-FLUX~\cite{ref4}} \\
\cmidrule(lr){2-3}\cmidrule(lr){4-5}
Setting & Native & RelGeo & Native & RelGeo \\
\midrule
SPair-71k & 57.63 & \rgGain{61.55}{3.92} & 66.88 & \rgGain{69.43}{2.55} \\
SOCO-CC & 60.57 & \rgGain{68.02}{7.45} & 71.40 & \rgGain{76.38}{4.98} \\
SOCO-SOC & 48.89 & \rgGain{58.02}{9.13} & 56.58 & \rgGain{64.52}{7.94} \\
Cross-SOC & 42.67 & \rgGain{48.82}{6.15} & 50.84 & \rgGain{56.32}{5.48} \\
AP-10K & 47.80 & \rgGain{51.42}{3.62} & 52.55 & \rgGain{56.46}{3.91} \\
\bottomrule
\end{tabular}
\end{table}

\vspace{2pt}\noindent\textbf{Performance across backbones and settings.}\enspace
RelGeo improves all ten controlled evaluations by an average of 5.51 PCK@0.10 points (Table~\ref{tab:1}). The largest gains occur on SOCO-SOC (+9.13 for SD2.1 and +7.94 for FLUX), where correspondence identity is explicitly evaluated. We next examine candidate selection.

\begin{table*}[!t]
\centering
\caption{SPair-71k per-class PCK@0.10 (\%). F: no additional semantic-correspondence training; A: keypoint-free adaptation; S: manual-keypoint supervision. Published scores retain their source evaluation and aggregation: DiTF~\cite{ref4}, following GeoAware~\cite{ref6}, reports supervised methods per image and unsupervised methods per point. RelGeo uses class-macro per-point PCK. $\dagger$: independent DINO appearance fusion; $\ddagger$: keypoint-guided validation selection. Bold/underline mark best/second within each displayed block; gray denotes RelGeo.}
\label{tab:published_spair_full}
\begingroup
\footnotesize
\setlength{\tabcolsep}{1.5pt}
\renewcommand{\arraystretch}{0.86}
\begin{tabular}{@{}ll*{19}{r}@{}}
\toprule
 & \textbf{Method} & \textbf{Aero} & \textbf{Bike} & \textbf{Bird} & \textbf{Boat} & \textbf{Bott} & \textbf{Bus} & \textbf{Car} & \textbf{Cat} & \textbf{Chai} & \textbf{Cow} & \textbf{Dog} & \textbf{Hors} & \textbf{Moto} & \textbf{Pers} & \textbf{Plant} & \textbf{Shee} & \textbf{Trai} & \textbf{TV} & \textbf{Overall} \\
\midrule
S & PMNC-best~\cite{ref43} & 54.1 & 35.9 & 74.9 & 36.5 & 42.1 & 48.8 & 40.0 & 72.8 & 21.1 & 67.6 & 58.1 & 50.5 & 40.1 & 54.1 & 43.3 & 35.7 & 74.5 & 59.9 & 50.4 \\
S & SCorrSAN~\cite{ref42} & 57.1 & 40.3 & 78.3 & 38.1 & 51.8 & 57.8 & 47.1 & 67.9 & 25.2 & 71.3 & 63.9 & 49.3 & 45.3 & 49.8 & 48.8 & 40.3 & 77.7 & \underline{69.7} & 55.3 \\
S & CATs++~\cite{ref16} & 60.6 & 46.9 & 82.5 & \underline{41.6} & \underline{56.8} & 64.9 & 50.4 & 72.8 & 29.2 & 75.8 & \underline{65.4} & 62.5 & 50.9 & 56.1 & 54.8 & 48.2 & 80.9 & \textbf{74.9} & 59.8 \\
S & DHF~\cite{ref2} & \underline{74.0} & \underline{61.0} & \underline{87.2} & 40.7 & 47.8 & \underline{70.0} & \underline{74.4} & \underline{80.9} & \underline{38.5} & \underline{76.1} & 60.9 & \underline{66.8} & \underline{66.6} & \underline{70.3} & \underline{58.0} & \underline{54.3} & \underline{87.4} & 60.3 & \underline{64.9} \\
S & SD+DINO (S)~\cite{ref5} & \textbf{81.2} & \textbf{66.9} & \textbf{91.6} & \textbf{61.4} & \textbf{57.4} & \textbf{85.3} & \textbf{83.1} & \textbf{90.8} & \textbf{54.5} & \textbf{88.5} & \textbf{75.1} & \textbf{80.2} & \textbf{71.9} & \textbf{77.9} & \textbf{60.7} & \textbf{68.9} & \textbf{92.4} & 65.8 & \textbf{74.6} \\
\midrule
A & ASIC~\cite{ref40} & 57.9 & 25.2 & 68.1 & 24.7 & 35.4 & 28.4 & 30.9 & 54.8 & 21.6 & 45.0 & 47.2 & 39.9 & 26.2 & 48.8 & 14.5 & 24.5 & 49.0 & 24.6 & 36.9 \\
F & SCOT$^{\ddagger}$~\cite{ref13} & 34.9 & 20.7 & 63.8 & 21.1 & 43.5 & 27.3 & 21.3 & 63.1 & 20.0 & 42.9 & 42.5 & 31.1 & 29.8 & 35.0 & 27.7 & 24.4 & 48.4 & 40.8 & 35.6 \\
F & DINOv2+NN~\cite{ref27} & 72.7 & 62.0 & 85.2 & 41.3 & 40.3 & 52.3 & 51.5 & 71.1 & 36.2 & 67.1 & 64.6 & 67.6 & 61.0 & 68.2 & 30.7 & 62.0 & 54.3 & 24.2 & 55.6 \\
F & DIFT~\cite{ref1} & 63.5 & 54.5 & 80.8 & 34.5 & 46.2 & 52.7 & 48.3 & 77.7 & 39.0 & 76.0 & 54.9 & 61.3 & 53.3 & 46.0 & 57.8 & 57.1 & 71.1 & 63.4 & 57.7 \\
F & DiTF-SD3.5~\cite{ref4} & 66.5 & 56.8 & 86.3 & 40.1 & 51.3 & 58.6 & 58.8 & 81.3 & \textbf{47.8} & 79.6 & 60.2 & 68.5 & 65.7 & 73.5 & \textbf{64.4} & 67.8 & 69.7 & \textbf{66.5} & 64.6 \\
F & DiTF-FLUX~\cite{ref4} & \underline{74.3} & \underline{65.0} & \underline{88.1} & \textbf{48.1} & \underline{53.2} & 60.7 & \underline{60.7} & \underline{84.9} & 42.4 & \underline{82.8} & \underline{68.4} & \underline{72.1} & \underline{70.9} & \textbf{74.2} & \underline{62.1} & \textbf{72.6} & 66.0 & 60.3 & \underline{67.1} \\
\rowcolor{black!5}
F & \textbf{RelGeo-SD2.1} & 68.3 & 58.7 & 81.9 & 36.2 & 48.2 & \underline{63.1} & 55.0 & 79.2 & 42.9 & 79.4 & 57.9 & 65.1 & 59.0 & 55.0 & 56.2 & 58.9 & \underline{76.3} & \textbf{66.5} & 61.5 \\
\rowcolor{black!5}
F & \textbf{RelGeo-FLUX} & \textbf{75.1} & \textbf{66.2} & \textbf{88.6} & \underline{47.3} & \textbf{54.6} & \textbf{69.9} & \textbf{64.5} & \textbf{87.3} & \underline{47.5} & \textbf{84.5} & \textbf{70.1} & \textbf{73.3} & \textbf{72.4} & \underline{73.8} & 59.3 & \underline{72.0} & \textbf{77.4} & \underline{65.9} & \textbf{69.4} \\
\midrule
A & Shape-of-You$^{\dagger}$~\cite{ref7} & 73.5 & 66.7 & 89.9 & 40.8 & \underline{58.0} & 55.4 & 51.1 & 84.8 & 52.1 & 81.0 & 71.0 & \underline{75.3} & 64.1 & 71.8 & 62.1 & 70.2 & 70.2 & 64.4 & 67.9 \\
F & SD+DINO$^{\dagger}$~\cite{ref5} & 73.0 & 64.1 & 86.4 & 40.7 & 52.9 & 55.0 & 53.8 & 78.6 & 45.5 & 77.3 & 64.7 & 69.7 & 63.3 & 69.2 & 58.4 & 67.6 & 66.2 & 53.5 & 64.0 \\
F & Fuse2Match$^{\dagger}$~\cite{ref48} & 75.1 & 66.2 & 87.7 & 43.4 & 55.3 & 61.4 & 57.1 & 81.1 & 51.9 & 82.1 & 67.1 & 71.6 & 67.9 & 68.8 & \textbf{66.2} & 69.6 & 75.4 & \textbf{69.9} & 68.8 \\
F & GeoAware-SC$^{\dagger}$~\cite{ref6} & \underline{78.0} & 66.4 & \underline{90.2} & 44.5 & \textbf{60.1} & 66.6 & 60.8 & 82.7 & \underline{53.2} & 82.3 & 69.5 & 75.1 & 66.1 & 71.7 & 58.9 & 71.6 & \underline{83.8} & 55.5 & 69.6 \\
F & DiTF-SD3.5$^{\dagger}$~\cite{ref4} & \textbf{79.1} & \textbf{67.8} & \textbf{90.6} & 48.3 & 56.1 & \underline{69.2} & \textbf{65.0} & \underline{85.7} & \textbf{59.4} & \textbf{84.6} & \textbf{72.6} & \textbf{76.7} & 69.9 & \underline{75.5} & 62.1 & \textbf{74.4} & \textbf{85.6} & 59.6 & \textbf{72.2} \\
F & DiTF-FLUX$^{\dagger}$~\cite{ref4} & 76.3 & \textbf{67.8} & 88.5 & \textbf{50.8} & 55.8 & 60.9 & 60.8 & 82.9 & 47.1 & 81.3 & 70.5 & 72.3 & \underline{70.0} & \textbf{76.2} & 62.6 & 72.5 & 64.8 & 56.5 & 67.6 \\
\rowcolor{black!5}
F & \textbf{RelGeo-SD2.1$^{\dagger}$} & 73.8 & 64.2 & 86.7 & 41.1 & 54.5 & 67.6 & 61.6 & 83.3 & \underline{53.2} & 82.6 & 66.8 & 72.7 & 64.5 & 68.2 & \underline{63.5} & 68.2 & 79.0 & \underline{69.8} & 67.8 \\
\rowcolor{black!5}
F & \textbf{RelGeo-FLUX$^{\dagger}$} & 75.8 & \underline{67.5} & 88.3 & \underline{48.6} & 57.8 & \textbf{70.8} & \underline{64.6} & \textbf{87.3} & 51.4 & \underline{84.5} & \underline{71.3} & 73.8 & \textbf{70.4} & 74.6 & 63.4 & \underline{72.6} & 76.2 & 65.0 & \underline{70.2} \\
\bottomrule
\end{tabular}
\endgroup
\end{table*}

\vspace{2pt}\noindent\textbf{Comparison on SPair-71k.}\enspace
On SPair-71k, RelGeo-FLUX reaches 69.43\% without independent DINO fusion, compared with the reported 67.1\% of DiTF-FLUX (Table~\ref{tab:published_spair_full}). Under matched FLUX extraction, RelGeo improves Native from 66.88\% to 69.43\%. Independent DINO fusion yields 67.8\% for SD2.1 and 70.22\% for FLUX.

\begin{table}[!htbp]
\centering
\caption{SOCO-SOC PCK@0.10 (\%). F/A/S and $\dagger$: Table~\ref{tab:published_spair_full}. Published baselines: SOCO~\cite{ref8}, Table 4.}
\label{tab:3}
\small
\renewcommand{\arraystretch}{0.95}
\begin{tabular}{@{}llr@{}}
\toprule
Method & Setting & SOC \\
\midrule
SD (GeoAware-SC) & F & 47.7 \\
SD + DINO (GeoAware-SC) & F & 62.9 \\
CleanDIFT + DINO & F & 63.4 \\
DIY-SC & A & 69.2 \\
TLR (DINO) & S & 72.7 \\
TLR (SD + DINO) & S & 72.9 \\
\midrule
RelGeo-SD2.1 & F & 58.02 \\
\textbf{RelGeo-FLUX} & F & \textbf{64.52} \\
RelGeo-SD2.1$^{\dagger}$ & F & 66.61 \\
\textbf{RelGeo-FLUX$^{\dagger}$} & F & \textbf{68.39} \\
\bottomrule
\end{tabular}
\end{table}

\vspace{2pt}\noindent\textbf{Comparison on SOCO-SOC.}\enspace
On SOCO-SOC, RelGeo-FLUX reaches 64.52\% without independent DINO fusion, compared with 62.9\% for GeoAware-SC (SD+DINO) and 63.4\% for CleanDIFT+DINO. Independent DINO fusion raises RelGeo-FLUX to 68.39\% (Table~\ref{tab:3}).

\begin{figure*}[!t]
\centering
\includegraphics[width=0.80\textwidth]{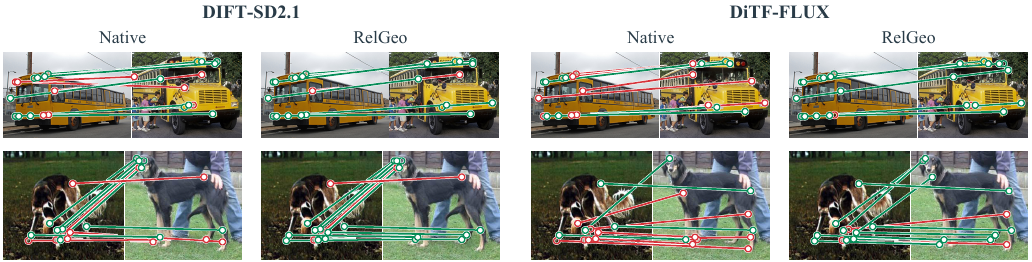}
\caption{Qualitative comparison. Green lines indicate correct matches; red lines indicate errors. RelGeo improves matching under rigid viewpoint and non-rigid pose variations through pair-conditioned relations. Please refer to Supplementary S9 for more results.}
\label{fig:qualitative}
\end{figure*}

\subsection{Calibration of Existing Candidates}
\label{sec:repair}

\begin{table}[!htbp]
\centering
\caption{Cross-SOC PCK@0.10 and break rates (\%): Native Top-5 (Fixed) versus full-domain (Full) readout. Break is the fraction of Native-correct predictions made incorrect.}
\label{tab:fixed_main}
\small
\renewcommand{\arraystretch}{0.95}
\begin{tabular}{@{}lrrrr@{}}
\toprule
 & \multicolumn{2}{c}{PCK $\uparrow$} & \multicolumn{2}{c}{Break $\downarrow$} \\
\cmidrule(lr){2-3}\cmidrule(lr){4-5}
Readout & SD2.1 & FLUX & SD2.1 & FLUX \\
\midrule
Native & 42.67 & 50.84 & 0.00 & 0.00 \\
RelGeo Fixed & \textbf{47.26} & \textbf{55.30} & 6.28 & 6.60 \\
RelGeo Full & \textbf{48.82} & \textbf{56.32} & 7.10 & 7.48 \\
\bottomrule
\end{tabular}
\end{table}

\vspace{2pt}\noindent\textbf{Fixed-candidate reranking.}\enspace
We rerank Native's original five NMS peaks, holding candidate positions and MR@5 coverage fixed. On Cross-SOC, SRC alone raises PCK from 42.67\% to 44.86\% for SD2.1 and from 50.84\% to 53.08\% for FLUX. Full RelGeo reaches 47.26\% and 55.30\%, gains of 4.59 and 4.46 points attributable to candidate selection (Table~\ref{tab:fixed_main}). Full-domain matching further increases PCK to 48.82\% and 56.32\%.

\vspace{2pt}\noindent\textbf{Correction of available candidates.}\enspace
Error-transition statistics further test the selection bottleneck identified in Sec.~\ref{sec:selection_failures}. On SOCO-SOC, RelGeo corrects 38.02\%/35.09\% of Native errors with a valid Top-5 alternative, compared with 9.54\%/10.06\% of errors without one (SD2.1/FLUX). Meanwhile, 4.67\%/4.88\% of Native-correct predictions become incorrect (Supplementary Fig.~S1). The higher conditional repair rate for candidate-available errors connects RelGeo's corrections to candidates already retrieved by Native.

\vspace{2pt}\noindent\textbf{Correspondence identity.}\enspace
RoleAcc, the class-macro rate of the designated match outranking other same-concept annotations, rises on SOCO-SOC from 63.17\% to 66.91\% for SD2.1 and from 59.66\% to 66.29\% for FLUX (Supplementary S7). The gap between backbones narrows from 3.51 to 0.62 points, directly addressing FLUX's weaker same-concept ranking in Sec.~\ref{sec:selection_failures}.

\subsection{Contribution of Relational Sources}

\begin{table}[!htbp]
\centering
\caption{Source ablation (PCK@0.10, \%). All variants retain semantic descriptors and cosine readout. Cross: Cross-SOC; AP: AP-10K. WIC uses pretrained transport.}
\label{tab:ablation}
\begingroup
\small
\setlength{\tabcolsep}{2.8pt}
\renewcommand{\arraystretch}{0.95}
\begin{tabular}{@{}llrrrrr@{}}
\toprule
Backbone & Variant & SPair & CC & SOC & Cross & AP \\
\midrule
SD2.1 & Native & 57.63 & 60.57 & 48.89 & 42.67 & 47.80 \\
 & +SRC & 58.29 & 63.27 & 51.07 & 45.18 & 48.59 \\
 & +WIC & 60.59 & 66.31 & 56.64 & 46.87 & 50.85 \\
 & \textbf{RelGeo} & \textbf{61.55} & \textbf{68.02} & \textbf{58.02} & \textbf{48.82} & \textbf{51.42} \\
\midrule
FLUX & Native & 66.88 & 71.40 & 56.58 & 50.84 & 52.55 \\
 & +SRC & 67.67 & 73.48 & 58.89 & 53.46 & 54.26 \\
 & +WIC & 68.85 & 75.11 & 63.48 & 54.49 & 55.40 \\
 & \textbf{RelGeo} & \textbf{69.43} & \textbf{76.38} & \textbf{64.52} & \textbf{56.32} & \textbf{56.46} \\
\bottomrule
\end{tabular}
\endgroup
\end{table}

\begin{figure}[!htbp]
\centering
\includegraphics[width=0.9\linewidth]{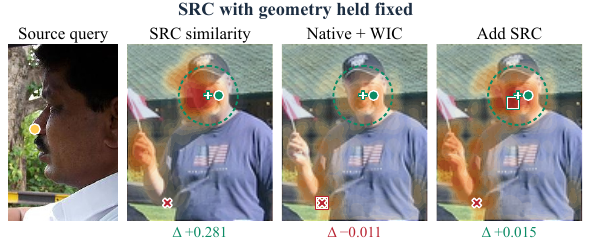}
\caption{SRC complements geometric transport. For a FLUX query, adding SRC to Native+WIC changes the peak order and places the full-domain maximum within the valid PCK region. WIC is fixed; fused responses share a scale.}
\label{fig:relational_response}
\end{figure}

\vspace{2pt}\noindent\textbf{Independent contributions of SRC and WIC.}\enspace
Single-source ablations test the contribution of each relational origin. SRC alone improves all ten settings by 1.84 PCK@0.10 points on average, showing that reciprocal frozen Q/K associations provide useful discrimination beyond the original descriptors (Table~\ref{tab:ablation}). WIC alone gains 4.28 points, consistent with geometric transport's stronger raw preference for valid candidates in Sec.~\ref{sec:raw_evidence}. Both relation sources thus improve Native independently.

\vspace{2pt}\noindent\textbf{Complementarity of internal and external relations.}\enspace
The central complementarity test is whether internal associations remain useful once external transport is present. Adding SRC to Native+WIC improves all ten settings by a further 1.24 points on average. This is consistent with Sec.~\ref{sec:raw_evidence}, where Q/K sometimes favors a valid candidate rejected by transport. With WIC fixed, Fig.~\ref{fig:relational_response} shows SRC promoting a valid peak; the branches also correct partly different SOCO-SOC queries (Supplementary S6).

Supplementary S6 also compares WIC with direct RoMa prediction and forward Gaussian fusion.

\subsection{Dependence on Current-Pair Relations}
\label{sec:pair_dependence}

\vspace{2pt}\noindent\textbf{From raw preference to calibrated ordering.}\enspace
On SOCO-SOC, when raw Q/K favors the valid candidate, SRC reverses the incorrect pairwise order in 31.97\%/32.79\% of eligible SD2.1/FLUX cases, versus 2.58\%/7.02\% when Q/K favors the incorrect candidate (Supplementary S4). Similar trends on AP-10K further connect raw associations with SRC's calibrated ordering.

\vspace{2pt}\noindent\textbf{Dependence on current-pair alignment.}\enspace
We keep semantic descriptors and cosine readout fixed, replacing relational coordinates with dimension-matched random SRC features or same-class relations from another image pair.

\begin{table}[!htbp]
\centering
\caption{SOCO-SOC relation controls (PCK@0.10, \%). Random: dimension-matched SRC; Swap/Pair: same-class mismatched/aligned relations. WIC swaps include reliability.}
\label{tab:4}
\begingroup
\small
\setlength{\tabcolsep}{2.2pt}
\renewcommand{\arraystretch}{0.95}
\begin{tabular}{@{}lrrrrrr@{}}
\toprule
 & & \multicolumn{3}{c}{SRC} & \multicolumn{2}{c}{WIC} \\
\cmidrule(lr){3-5}\cmidrule(lr){6-7}
Backbone & Native & Random & Swap & Pair & Swap & Pair \\
\midrule
SD2.1 & 48.89 & 48.79 & 47.50 & \textbf{51.07} & 46.65 & \textbf{56.64} \\
FLUX & 56.58 & 56.38 & 54.90 & \textbf{58.89} & 53.53 & \textbf{63.48} \\
\bottomrule
\end{tabular}
\endgroup
\end{table}

Aligned SRC reaches 51.07\%/58.89\% on SOCO-SOC (SD2.1/FLUX), versus 47.50\%/54.90\% with mismatched relations and 48.79\%/56.38\% with random features (Table~\ref{tab:4}). Aligned WIC exceeds mismatched WIC by 9.99/9.95 points; on AP-10K, replacing WIC and its reliability information with mismatched relations reduces PCK by 7.24/6.27 points (Supplementary S6). These controls establish the importance of relational alignment with the current image pair beyond added feature dimensions.

\section{Conclusion}
\label{sec:conclusion}
RelGeo calibrates frozen diffusion correspondence by expressing reciprocal Query/Key associations and pretrained bidirectional transport as aligned source--target coordinates, while retaining full-domain cosine nearest-neighbor matching. Our analysis shows that stronger descriptors improve valid-candidate retrieval more than final selection, and fixed-candidate reranking demonstrates that pair-conditioned evidence can correct mistakes among already retrieved alternatives. These findings reveal an additional source of correspondence capability in pretrained models: relational evidence that becomes useful when incorporated into the matching decision.

\label{main-content-end}
\typeout{RELGEOCONTENTPAGES=\thepage}
\clearpage
{\small\bibliographystyle{ieeenat_fullname}
\bibliography{main}
}
\clearpage
\renewcommand{\maketitlesupplementary}{%
  \begingroup
  \centering
  \large
  \textbf{\thetitle}\par
  \vspace{0.5em}Supplementary Material\par
  \vspace{0.7em}
  \endgroup
}

\maketitlesupplementary
\setcounter{section}{0}
\setcounter{figure}{0}
\setcounter{table}{0}
\setcounter{equation}{0}
\renewcommand{\thesection}{S\arabic{section}}
\renewcommand{\thefigure}{S\arabic{figure}}
\renewcommand{\thetable}{S\arabic{table}}
\renewcommand{\theequation}{S\arabic{equation}}
\renewcommand{\theHsection}{S\arabic{section}}
\renewcommand{\theHsubsection}{S\arabic{section}.\arabic{subsection}}
\renewcommand{\theHfigure}{S\arabic{figure}}
\renewcommand{\theHtable}{S\arabic{table}}
\renewcommand{\theHequation}{S\arabic{equation}}
\setlength{\textfloatsep}{12pt plus 2pt minus 2pt}
\makeatletter\setlength{\@fptop}{5pt}\setlength{\@fpsep}{10pt}\setlength{\@fpbot}{0pt}\makeatother
\setlength{\floatsep}{10pt plus 2pt minus 2pt}
\setlength{\intextsep}{10pt plus 2pt minus 2pt}
\begingroup
\small
\noindent\textbf{Contents}\par\vspace{.3em}
\newcommand{\suppcontentsentry}[3]{%
  \noindent\parbox[t]{.89\columnwidth}{\raggedright\hyperref[#3]{\textcolor{cvprblue}{#1.\enspace #2}}}%
  \hfill\hyperref[#3]{\pageref*{#3}}\par\vspace{.12em}}
\suppcontentsentry{S1}{Evaluation Protocols}{sec:supp1}
\suppcontentsentry{S2}{Implementation Details}{sec:supp2}
\suppcontentsentry{S3}{Candidate Availability and Competition}{sec:supp3}
\suppcontentsentry{S4}{Raw Relational Evidence}{sec:supp4}
\suppcontentsentry{S5}{Candidate Reranking and Correction}{sec:supp5}
\suppcontentsentry{S6}{Pair-Specific Relations and Source Complementarity}{sec:supp6}
\suppcontentsentry{S7}{Correspondence Identity Analysis}{sec:supp7}
\suppcontentsentry{S8}{Published Comparisons and Implementation Scope}{sec:supp8}
\suppcontentsentry{S9}{Additional Qualitative Correspondence Results}{sec:supp9}
\endgroup
\vspace{.8em}
\section{Evaluation Protocols}
\label{sec:supp1}

We evaluate frozen DIFT-SD2.1 and DiTF-FLUX on SPair-71k, SOCO-CC, SOCO-SOC, Cross-SOC, and AP-10K using each benchmark's PCK@0.10 normalization and coordinate adapter. SPair-71k contains 18 classes, 12,234 test pairs, and 88,328 queries. SOCO-SOC contains 100 classes, 20,000 image pairs, and 149,646 valid queries per pipeline; SOCO-CC uses 180,984 concept queries on those pairs, while Cross-SOC contains 122,540 queries over 100 classes and 20,000 pairs. AP-10K intra-species matching uses the evaluated test set of 50 classes, 20,439 pairs, and 150,504 valid queries.

\paragraph{Statistics and aggregation.}
Let $E$ denote a Native Top-1 error and $A_K$ the presence of a PCK-valid location among Native's first $K$ spatial peaks. MR@$K$ measures $A_K$; SF@5 measures $A_5$ conditional on $E$. Available-candidate error is the joint event $E\cap A_5$. Repair conditions on Native errors and break on Native-correct queries. Conditional rates use their own nonempty support before the averaging steps in Table~\ref{tab:aggregation}; pooled counts report support rather than macro weights. Annotation labels enter only post-prediction diagnostics, leaving feature extraction and candidate selection unchanged.

\begin{table}[htbp]
\centering
\caption{Aggregation used throughout the paper. Q: average eligible queries; P: average nonempty image pairs; C: average classes equally. Each arrow denotes a successive averaging step.}
\label{tab:aggregation}
\footnotesize
\resizebox{\columnwidth}{!}{%
\begin{tabular}{@{}ll@{}}
\toprule
Statistic & Aggregation \\
\midrule
SPair/AP PCK and candidate coverage & Q within class $\to$ C \\
SOCO PCK and candidate coverage & Q within pair $\to$ P within class $\to$ C \\
Raw preferences and fixed-candidate outcomes & SOCO: Q $\to$ P $\to$ C; AP: Q $\to$ C \\
SOC conditional repair and RoleAcc & Q within class $\to$ C \\
Branch success sets & Pooled queries within each success group \\
\bottomrule
\end{tabular}
}
\end{table}

\paragraph{Extraction and controlled comparisons.}
Within each execution path, Native and calibrated branches share inputs, extraction presets, queries, and cosine readout. SOCO uses empty prompts for both backbones. On AP-10K, FLUX uses an empty caption and SD2.1 uses \texttt{a photo of a \{category\}}. FLUX-SPair uses the supplied detailed captions, channel discard, eight sampling members, and the upstream unseeded extraction mode; SD2.1-SPair uses the DIFT evaluator's category conditioning. Both SPair evaluators report class-macro mean per-point PCK: within each of the 18 categories, correct valid keypoints are divided by all valid keypoints, and the resulting category scores are averaged equally. The SD2.1 evaluator names this field \texttt{per\_point\_category\_macro}, while the FLUX evaluator uses \texttt{Mean/per\_point}. The two names denote the same aggregation formula, not a pooled micro average. Query-level controls retain their recorded Native states; backbone comparisons also involve different extraction configurations.

\paragraph{Pretraining and supervision.}
In published comparisons, F denotes no additional semantic-correspondence training, A keypoint-free correspondence adaptation, and S manual-keypoint supervision. Frozen RoMa uses geometric supervision and DINOv2-L/14; the $\dagger$ appearance branch separately uses DINOv2-B/14. These labels describe correspondence training; pretraining resources still differ. Published scores retain their source readout and aggregation, with SPair and appearance-fusion configurations detailed in Sec.~\ref{sec:supp8}.

\paragraph{Uncertainty.}
The aligned SOC/AP raw-preference analyses and SOC multiplicity contrast use 5,000 class resamples with seed 20261026; paired contrasts resample the same classes together. The Cross-SOC multiplicity extension uses its recorded paired-class intervals. All intervals describe class variation under fixed recorded predictions. Differences are computed before rounding.

\section{Implementation Details}
\label{sec:supp2}

\subsection{Frozen Descriptors and Internal Associations}
Table~\ref{tab:supp1} lists the fixed extraction settings and SRC dimensions for each execution path. FLUX uses DiTF's \texttt{flux-dev} configuration, corresponding to FLUX.1-dev~\cite{ref58}. SD2.1 retains pair-based forwards and noise; FLUX-SOCO reuses losslessly cached first-use image states and their sampling inputs.

\begin{table}[htbp]
\centering
\footnotesize\setlength{\tabcolsep}{2.75pt}
\caption{Frozen extraction settings. $E$: diffusion sampling members; $r$: SRC dimension. The fusion weight is $\lambda=0.35$. FLUX input sizes preserve the extraction adapter's aspect-ratio convention.}
\label{tab:supp1}
\begin{tabular}{@{}lllrlrr@{}}
\toprule
Backbone & Dataset & Input & $t$ & Descriptor layer & $E$ & $r$ \\
\midrule
SD2.1 & SPair & $768\times768$ & 261 & DIFT layer 1 & 8 & 128 \\
 & SOCO/AP & $800\times800$ & 250 & DIFT layer 1 & 1 & 128 \\
FLUX & SPair & 640 & 260 & DiTF block 28 & 8 & 96 \\
 & SOCO/AP & 800 & 260 & DiTF block 28 & 8 & 96 \\
\bottomrule
\end{tabular}
\end{table}

SD2.1 captures the original \texttt{to\_q}/\texttt{to\_k} projections at the last attention module in U-Net \texttt{up\_blocks[1]}, before its upsampler: \texttt{transformer\_blocks[0].attn1}. FLUX uses \texttt{single\_blocks[8]} (global zero-based block index 27), preceding descriptor index 28. Its cross-image projections are taken after Q/K normalization and before rotary positional encoding; the original-context term uses image queries and image/text keys after rotary encoding. All attention heads are retained.

For each head and matched sampling member, let
\[
L_{s\to t}(a,b)=Q_s(a)^\top K_t(b)/\sqrt{d_h},
\]
where $a,b$ index image tokens and $d_h$ is the head dimension. Let $b_s(a)$ be the log-sum-exp of the source query in its original attention context. The cross-image probability is
\begin{equation}
\Pi_{s\to t}(a,b)=
\frac{\exp L_{s\to t}(a,b)}{\exp b_s(a)+\sum_{b'\in\mathcal T_t}\exp L_{s\to t}(a,b')}.
\label{eq:supp2}
\end{equation}
The denominator retains original-context mass alongside cross-image candidates, so cross-image row sums need not equal one. We average normalized probabilities over the $EH$ head/member combinations, pairing source and target by ensemble index, then construct reciprocal $M$ as in main-paper Eq.~\eqref{eq:2}. Raw Q/K states are not averaged before normalization.

\subsection{Relational Kernel and Spectral Coordinates}
SD2.1 uses $B=M$; FLUX uses $B=M\odot H_1(M)$. With row, column, and global means $\mu_{\rm row},\mu_{\rm col},\mu$, the fixed kernel paths are:
\par\medskip\noindent\begin{minipage}{\linewidth}
\captionsetup{hypcap=false}
\centering
\captionof{table}{SRC kernel paths. Local support is disabled for SD2.1 and has radius one for FLUX.}
\label{tab:supp2}
\resizebox{\columnwidth}{!}{\begin{tabular}{@{}ll@{}}
\toprule
Execution path & $K^{\rm SRC}$ \\
\midrule
SD2.1 / SOCO, AP-10K & $B-\mu_{\rm row}-\mu_{\rm col}+\mu$ \\
SD2.1 / SPair; FLUX / all settings & $B-\mu_{\rm row}-\mu_{\rm col}+2\mu$ \\
\bottomrule
\end{tabular}}
\end{minipage}\par\medskip
Writing $J_i=I-\mathbf1_i\mathbf1_i^\top/n_i$, the two paths are $J_sBJ_t$ and $J_sBJ_t+\mu\mathbf1_s\mathbf1_t^\top$. The added rank-one mode is orthogonal to the centered term; when retained, it occupies a coordinate and changes subsequent token normalization.

For FLUX, source offsets $\delta$ are the eight nonzero positions in $\{-1,0,1\}^2$. Target offsets $\delta_t$ scale each component by the corresponding grid-size ratio and round half-integers to even integers. Let $V(a,b)$ contain offsets valid on both grids, $u(a')=\max_jM(a',j)$, and $v(b')=\max_iM(i,b')$. Then
\begin{equation}
\begin{aligned}
H_1(a,b)&=\frac{1}{\max(1,|V(a,b)|)}
\sum_{\delta\in V(a,b)}\frac{M_{a+\delta,b+\delta_t}}{\sqrt{u_\delta v_\delta}},\\
u_\delta&=\max(u(a+\delta),10^{-12}),\\
v_\delta&=\max(v(b+\delta_t),10^{-12}).
\end{aligned}
\label{eq:supp1}
\end{equation}
Boundary masks remove invalid offsets, with $H_1=0$ when none remain. Clamped indices provide safe reads without replicating boundary contributions.

The leading-subspace solver \texttt{r64\_i2} adds 64 sampling vectors to the output rank, performs two power iterations with local seed 2027, and computes bilateral factors of the projected matrix. SRC uses $Z_s=U_r\Sigma_r^{1/2}$ and $Z_t=V_r\Sigma_r^{1/2}$ with shared component order and scale. Source--target inner products reconstruct the low-rank kernel on the construction grids; FP16 casting, nearest-neighbor mapping to descriptor grids, and per-location normalization then produce the final fusion vectors.

\subsection{Shared Endpoint Reference and Reliability}
Frozen \texttt{roma\_outdoor}~\cite{ref9}, geometrically pretrained on MegaDepth with DINOv2-L/14, uses coarse/refinement resolutions 560/864 and FP16 mixed precision. SOCO-CC/SOC, Cross-SOC, and AP-10K reuse backbone-shared FP16 fields on fixed $50\times50$ grids. SPair uses separate caches: FP16 $48\times48$ fields for SD2.1 and FP32 aspect-ratio-preserving fields for FLUX, including $27\times40$ and $30\times40$ grids. Each cache covers the complete evaluated split in Sec.~\ref{sec:supp1}.

The source and target endpoint pairs $p_s,p_t$ in main-paper Eq.~\eqref{eq:4} share the same source--target order. Define the unit-normalized encoding
\[
\phi(v)=\mathcal N\!\left([2v-1,\{\sin(\pi f v),\cos(\pi f v)\}_{f\in\{1,2,4\}}]\right).
\]
The 4D pair and 2D target-domain projection produce 28 and 14 channels, respectively. Their independently normalized vectors receive equal energy, giving the 42-channel WIC in main-paper Eq.~\eqref{eq:wic}. For a source query and target candidate, the endpoint contribution is
\begin{equation}
\begin{aligned}
&\langle I_s^{\rm WIC}(q),I_t^{\rm WIC}(y)\rangle\\[-1pt]
&=\frac{\rho_s(q)\rho_t(y)}{2}\Bigl[\\[-2pt]
&\quad\langle\phi(p_s(q)),\phi(p_t(y))\rangle\\[-2pt]
&\quad+\langle\phi(W_{s\to t}(q)),\phi(g_t(y))\rangle\Bigr].
\end{aligned}
\label{eq:endpoint_kernel}
\end{equation}
The first term compares locations in the joint source--target domain, and the second compares their target-domain projections. When $W_{s\to t}(q)=g_t(y)$ and $W_{t\to s}(y)=g_s(q)$, both unit encodings coincide and the contribution reaches $\rho_s(q)\rho_t(y)$ on the construction grid. WIC thus incorporates agreement between the two directional endpoint hypotheses into the comparison alongside semantic similarity.

We set nonfinite sampled RoMa confidences to zero and clamp confidence values to be nonnegative, giving $c_i$. The median positive confidence $m_i$ has a lower bound of $10^{-12}$:
\[
S_i=\frac{c_i}{c_i+m_i},\qquad
\rho_i=\frac{\sqrt{S_i}}{1+e_i}.
\]
Support is zero if no positive confidence exists. A valid-content mask removes padding after computing the median. The source cycle residual is $W_{t\to s}(W_{s\to t}(q))-g_s(q)$; multiplying its two components by $W_s-1,H_s-1$ before taking the Euclidean norm gives $e_s$ in grid-cell units. The target definition is symmetric.

RoMa fields are sampled at descriptor-cell centers bilinearly with border padding and \texttt{align\_corners=False}. Center coordinates are converted through continuous pixel indices to $x/(W-1),y/(H-1)$ and clipped to $[0,1]$. WIC grids use equally spaced endpoints, so cycle reconstruction uses \texttt{align\_corners=}\allowbreak\texttt{True}. Nonfinite endpoints become 0.5 for NaN, 1 for positive infinity, and 0 for negative infinity. These conventions distinguish center-sampled fields from endpoint coordinates.

\subsection{Fusion and Benchmark Readout}
On construction grids, before quantization and continuous query sampling, write $a_i=\mathcal N(A_i)$ and $h_i=I_i^{\rm WIC}$. Then
\begin{equation}
\langle R_s,R_t\rangle=
\frac{\langle a_s,a_t\rangle+\langle h_s,h_t\rangle}
{\sqrt{(1+\|h_s\|_2^2)(1+\|h_t\|_2^2)}}.
\label{eq:supp5}
\end{equation}
Thus reliability affects both endpoint agreement and the final normalizer. For nonzero semantic and SRC components, their cosine terms receive weights $1/(1+\lambda^2)$ and $\lambda^2/(1+\lambda^2)$, respectively.

For SOCO/AP, FP16 descriptors are converted to FP32 and normalized per grid location. Source queries are bilinearly sampled with \texttt{align\_corners=}\allowbreak\texttt{True} and compared with every target grid cell. Predicted target indices are divided by width $W$, preserving the evaluator convention. For SPair, both maps are first lifted bilinearly to original image dimensions with \texttt{align\_corners=False}; source integer keypoints and target pixels are then normalized for cosine matching. The evaluator's quantization, interpolation, and normalization order is retained in every branch.

\section{Candidate Availability and Competition}
\label{sec:supp3}

Starting from Native Top-1, we repeatedly select the largest remaining response and suppress its square neighborhood to obtain five ordered spatial peaks. SOCO/AP use radius one on the target readout grid. SPair scales this radius with the descriptor-to-readout resolution ratio, with a minimum of one. Validity uses the original PCK threshold and coordinate adapter; the candidate set depends only on Native responses and includes Native Top-1.

\noindent\textbf{Visible same-concept alternatives.}\enspace Following SOCO~\cite{ref8}, $M_{\rm ann}$ counts target annotations sharing the source query's concept, including the specified counterpart. It is distinct from SRC's matrix $M$. The count describes visible annotations rather than response peaks or exhaustive symmetry labels. On 97 SOC classes supporting both $M_{\rm ann}=1$ and $M_{\rm ann}\ge2$, the groups contain 40,771 and 104,321 queries, identically for both backbones. Within each group, we average queries within pairs, pairs within classes, and these same classes equally.

\begin{table}[htbp]
\centering
\footnotesize\setlength{\tabcolsep}{2.75pt}
\caption{SOC competition on common class support (\%). $\Delta$: multiple minus one same-concept annotation, computed before rounding, with paired-class 95\% intervals.}
\label{tab:multiplicity_soc}
\resizebox{\columnwidth}{!}{%
\begin{tabular}{@{}llrrrr@{}}
\toprule
Backbone & Metric & $M_{\rm ann}=1$ & $M_{\rm ann}\ge2$ & $\Delta$ & 95\% CI \\
\midrule
SD2.1 & $P(E\cap A_5)$ & 18.24 & 26.33 & +8.09 & [6.40,9.83] \\
 & MR@5 & 75.56 & 71.06 & $-4.50$ & [$-6.74,-2.22$] \\
FLUX & $P(E\cap A_5)$ & 19.29 & 31.01 & +11.71 & [8.82,14.24] \\
 & MR@5 & 87.76 & 81.59 & $-6.18$ & [$-7.86,-4.60$] \\
\bottomrule
\end{tabular}
}
\end{table}

More annotated alternatives accompany more candidate-available errors despite lower coverage, motivating identity analysis alongside spatial availability (Sec.~\ref{sec:supp7}). The comparison uses common class support and distinct queries and image pairs, retaining each group's observed pose, scale, structure, and difficulty distributions. PCK validity measures spatial proximity, while annotated identity is evaluated separately.

\paragraph{Cross-category extension.}
Cross-SOC shows the same association: candidate-available error rises while coverage falls (Table~\ref{tab:cross_competition}). This comparison uses its own 97 common classes and 35,153/83,241 queries with one/multiple annotations, respectively, and retains pair--class averaging. Its class and query composition differs from SOC.

\begin{table}[htbp]
\centering
\footnotesize\setlength{\tabcolsep}{2.75pt}
\caption{Cross-SOC extension on common support within that setting (\%). Each metric uses the same 97 classes across multiplicity groups; differences use unrounded rates.}
\label{tab:cross_competition}
\begin{tabular}{@{}llrrr@{}}
\toprule
Backbone & Metric & $M_{\rm ann}=1$ & $M_{\rm ann}\ge2$ & Difference (points) \\
\midrule
SD2.1 & $P(E\cap A_5)$ & 20.43 & 27.90 & +7.46 \\
 & MR@5 & 69.36 & 67.10 & $-2.26$ \\
FLUX & $P(E\cap A_5)$ & 20.72 & 32.47 & +11.75 \\
 & MR@5 & 81.50 & 78.03 & $-3.47$ \\
\bottomrule
\end{tabular}
\end{table}

The available-error differences have paired-class 95\% intervals [5.79,9.02]/[9.61,13.66]. More alternatives also accompany lower repair and higher break rates. Across nonempty Cross-SOC strata, full RelGeo repair falls from 34.12/35.04\% at $M_{\rm ann}=1$ to 23.01/22.69\% at $M_{\rm ann}\ge3$, while break rises from 4.50/4.40\% to 10.51/9.93\%. Repair and break are evaluated on their respective eligible pair/class support within each stratum.

\section{Raw Relational Evidence}
\label{sec:supp4}

Internal associations and geometric endpoints provide distinct raw support on identical Native competitors. Positive raw preference accompanies more calibrated reversals.

\paragraph{Candidate comparisons.}
For each Native error with $A_5=1$, we compare the highest-ranked valid peak $y^+$ with the wrong Top-1 $y^-$. Both remain fixed throughout the analysis. On SOC/AP, Native, raw Q/K associations, and SRC share the same extraction; SD2.1 retains its original noise sequence with seed 8. RoMa endpoints are evaluated on these same competitors. The endpoint-to-RelGeo analysis uses a separate AP full-method evaluation, pairing raw preferences and calibrated outcomes within that evaluation.

Raw preferences describe the Native-defined competitors under asymmetric candidate roles, with no 50\% random-selection null. The post-hoc union (positive preference from either source) measures annotation-conditioned raw support, not realizable accuracy or calibrated-branch success.

\subsection{Internal Associations and External Endpoints}
Internal preference $I^+$ means $M(q,y^+)>M(q,y^-)$ before kernel processing, local support, spectral decomposition, or fusion. The raw association is mapped to descriptor grids and bilinearly sampled at the source query. External preference $G^+$ means
\begin{equation}
\begin{aligned}
d_{\rm end}(q,y^+)&<d_{\rm end}(q,y^-),\\
d_{\rm end}(q,y)&=\|W_{s\to t}(q)-g_t(y)\|_2^2\\
&\quad +\|W_{t\to s}(y)-g_s(q)\|_2^2.
\end{aligned}
\label{eq:raw_endpoint}
\end{equation}
The endpoint test precedes encoding and reliability weighting; minus signs denote the reverse preference. On $50\times50$ grids it uses endpoint coordinates $j/49$; candidate validity retains the official predictor's $j/50$. SOC preserves integer-resized queries and AP continuous centered padding.

\begin{table}[htbp]
\centering
\footnotesize\setlength{\tabcolsep}{2.75pt}
\caption{Aligned raw preferences on Native-defined competitors. Rates (\%) and 95\% class-bootstrap intervals use all eligible queries. $n$ is pooled support; eligible SOC pairs number 15,574/16,131 for SD2.1/FLUX.}
\label{tab:raw_preferences}
\resizebox{\columnwidth}{!}{%
\begin{tabular}{@{}llrrll@{}}
\toprule
Setting & Backbone & Classes & $n$ & Internal $I^+$ & Endpoint $G^+$ \\
\midrule
SOC & SD2.1 & 100 & 35,243 & 50.64 [49.27,51.98] & 72.84 [71.61,74.05] \\
 & FLUX & 100 & 39,030 & 39.80 [38.71,40.86] & 67.68 [65.58,69.75] \\
AP & SD2.1 & 50 & 34,460 & 43.64 [42.53,44.75] & 59.18 [57.12,61.14] \\
 & FLUX & 50 & 36,203 & 40.53 [39.30,41.71] & 56.92 [54.69,59.15] \\
\bottomrule
\end{tabular}
}
\end{table}

\begin{table}[htbp]
\centering
\footnotesize\setlength{\tabcolsep}{2.75pt}
\caption{Joint raw-source preferences on the same candidates (\%). The four cells sum to 100\% before rounding. Union reports at least one positive preference, with 95\% CI.}
\label{tab:paired_raw}
\resizebox{\columnwidth}{!}{%
\begin{tabular}{@{}llrrrrl@{}}
\toprule
Setting & Backbone & $I^+G^+$ & $I^+G^-$ & $I^-G^+$ & $I^-G^-$ & Union \\
\midrule
SOC & SD2.1 & 38.90 & 11.74 & 33.94 & 15.42 & 84.58 [83.69,85.42] \\
 & FLUX & 29.66 & 10.14 & 38.02 & 22.18 & 77.82 [76.29,79.33] \\
AP & SD2.1 & 28.43 & 15.21 & 30.75 & 25.61 & 74.39 [72.86,75.86] \\
 & FLUX & 26.18 & 14.35 & 30.75 & 28.72 & 71.28 [69.56,72.96] \\
\bottomrule
\end{tabular}
}
\end{table}

Each source favors the valid competitor on queries where the other favors the incorrect one (Table~\ref{tab:paired_raw}). Endpoints provide greater marginal support.

\subsection{Raw Preference and Calibrated Ordering}
The designated pair's strict reversal is $F_Z$ for SRC in internal records and $F_R$ for RelGeo in AP full-method records. Conditional rates use raw-sign support and the pair/class hierarchy in Sec.~\ref{sec:supp1}; another invalid peak can still rank first after pair reversal. Positive raw preference accompanies more calibrated reversals (Table~\ref{tab:ordering}).

\begin{table}[htbp]
\centering
\footnotesize\setlength{\tabcolsep}{2.75pt}
\caption{Ordering reversal conditional on raw preference (\%). Internal rows use $I^{\pm}$ and $F_Z$; endpoint rows use $G^{\pm}$ and $F_R$. $\Delta F$ is a paired-class contrast in points with 95\% CI.}
\label{tab:ordering}
\resizebox{\columnwidth}{!}{%
\begin{tabular}{@{}llllrrl@{}}
\toprule
Source & Setting & Backbone & Output & $F\mid+$ & $F\mid-$ & $\Delta F$ [95\% CI] \\
\midrule
Internal & SOC & SD2.1 & SRC & 31.97 & 2.58 & 29.39 [28.04,30.72] \\
 & SOC & FLUX & SRC & 32.79 & 7.02 & 25.78 [24.34,27.22] \\
 & AP & SD2.1 & SRC & 24.73 & 2.06 & 22.67 [21.37,23.92] \\
 & AP & FLUX & SRC & 28.70 & 6.19 & 22.51 [21.02,23.97] \\
Endpoint & AP & SD2.1 & RelGeo & 36.82 & 7.64 & --- \\
 & AP & FLUX & RelGeo & 43.83 & 9.14 & --- \\
\bottomrule
\end{tabular}
}
\end{table}

\section{Candidate Reranking and Correction}
\label{sec:supp5}

Fixed-candidate reranking improves accuracy while retaining most correct predictions. Fixed readout maximizes calibrated scores over Native's original five NMS peaks, whereas full readout searches the complete target domain. Both use the same queries, labels, and score maps within each recorded comparison. Repair conditions on $E\cap A_5$, absent-candidate repair on $E\cap\neg A_5$, and break on Native-correct queries. The strict two-peak flip between $y^+$ and $y^-$ in Sec.~\ref{sec:supp4} compares the designated competitors; fixed readout instead maximizes over all five peaks. SOCO uses pair--class averaging; AP pools within classes.

\begin{table}[htbp]
\centering
\footnotesize\setlength{\tabcolsep}{2.75pt}
\caption{Fixed/full readout on complete Cross-SOC and AP evaluations (\%). PCK uses all queries; break and repair use their respective Native-defined conditioning sets. Uniform gives its analytical expectation on the original five peaks.}
\label{tab:fixed_global}
\resizebox{\columnwidth}{!}{%
\begin{tabular}{@{}llrrrrrr@{}}
\toprule
 & & \multicolumn{2}{c}{PCK} & \multicolumn{2}{c}{Break} & \multicolumn{2}{c}{Repair} \\
\cmidrule(lr){3-4}\cmidrule(lr){5-6}\cmidrule(lr){7-8}
Setting & Readout & SD2.1 & FLUX & SD2.1 & FLUX & SD2.1 & FLUX \\
\midrule
Cross-SOC & Native & 42.67 & 50.84 & 0 & 0 & 0 & 0 \\
 & Uniform Top-5 & 28.41 & 30.55 & 52.92 & 57.39 & 27.63 & 28.31 \\
 & SRC fixed & 44.86 & 53.08 & 6.11 & 5.97 & 16.90 & 17.49 \\
 & SRC full & 45.18 & 53.46 & 6.57 & 6.37 & 17.49 & 18.07 \\
 & RelGeo fixed & 47.26 & 55.30 & 6.28 & 6.60 & 26.89 & 27.02 \\
 & RelGeo full & 48.82 & 56.32 & 7.10 & 7.48 & 28.45 & 27.94 \\
\midrule
AP & Native & 47.80 & 52.55 & 0 & 0 & 0 & 0 \\
 & Uniform Top-5 & 35.98 & 32.48 & 39.42 & 51.43 & 28.15 & 28.79 \\
 & RelGeo fixed & 50.32 & 55.60 & 5.26 & 5.87 & 22.06 & 26.21 \\
 & RelGeo full & 51.42 & 56.46 & 6.56 & 6.62 & 24.41 & 27.17 \\
\bottomrule
\end{tabular}
}
\end{table}

\paragraph{Reranking with retention.}
Fixed readout raises Cross-SOC PCK by 4.59/4.46 points and improves both AP pipelines; SRC alone also improves Cross-SOC without external transport. Uniform selection has expected correctness $v/5$ when $v$ peaks are valid. Although it repairs more candidate-available errors than restricted RelGeo, it breaks 39.42--57.39\% of Native-correct predictions and lowers overall PCK.

\begin{table}[htbp]
\centering
\footnotesize\setlength{\tabcolsep}{2.75pt}
\caption{Full-domain correction and strict two-peak reversal (\%). Available/absent refer to valid Native Top-5 coverage among Native errors. Flip compares only the designated valid peak with the wrong Top-1.}
\label{tab:supp5}
\begin{tabular}{@{}llrrrr@{}}
\toprule
Setting & Backbone & Available repair & Absent repair & Break & Flip \\
\midrule
Cross-SOC & SD2.1 & 28.45 & 5.58 & 7.10 & 29.92 \\
 & FLUX & 27.94 & 6.39 & 7.48 & 29.82 \\
AP & SD2.1 & 24.41 & 3.79 & 6.56 & 24.91 \\
 & FLUX & 27.17 & 4.25 & 6.62 & 29.06 \\
\bottomrule
\end{tabular}
\end{table}

\begin{sloppypar}
RelGeo reverses the designated Cross-SOC pair in 29.92\%/29.82\% of eligible errors, compared with SRC's 19.37\%/19.72\%. The separate internal SOC/AP records yield SRC fixed PCK of 50.77\%/58.53\% and 48.42\%/54.02\%, respectively. These internally matched outputs form a separate evaluation from the full-method AP records in Table~\ref{tab:fixed_global}.
\end{sloppypar}

\begin{figure}[tbp]
\centering
\includegraphics[width=\columnwidth]{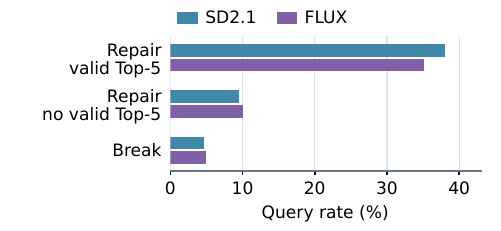}
\caption{SOCO-SOC repair and break rates. Repair is measured within each Native-error group. Break is measured on Native-correct queries.}
\label{fig:repair_identity}
\end{figure}

\paragraph{Full-domain localization.}
Full readout often recovers competitive regions while allowing nearby shifts. Fixed and full readouts each correct queries missed by the other: on AP candidate-available errors, their joint outcomes (both correct, fixed only, full only, neither) are $(19.76,2.30,4.65,73.29)\%$ for SD2.1 and $(24.54,1.67,2.63,71.16)\%$ for FLUX. The two readouts thus recover overlapping but distinct subsets of candidate-available errors.

\section{Pair-Specific Relations and Source Complementarity}
\label{sec:supp6}

The main-paper ablation (Table~\ref{tab:ablation}) shows that SRC and WIC each improve Native across all ten evaluations, and adding SRC to Native+WIC yields further gains. We examine the contribution of geometric endpoints, the dependence on correct pair alignment, and the overlap between corrections made by the two sources.

\subsection{Direct Transport and Gaussian Fusion}
Direct RoMa reads the source query's endpoint from the fields used by WIC. Gaussian warp fusion adds a forward-endpoint prior to the Native response:
\begin{equation}
s_{\mathrm G}(q,y)=s_D(q,y)+0.5\exp\!\left(
-\frac{\|u_t(y)-w(q)\|_2^2}{2(0.05)^2}
\right).
\label{eq:gaussian_fusion}
\end{equation}
Here $w(q)$ is the sampled RoMa endpoint and $u_t(y)$ is the target coordinate in the pipeline's normalized readout grid. The full-domain maximum gives the prediction. All methods retain their benchmark's query sampling and PCK aggregation; SPair uses each pipeline's field mapping from Sec.~\ref{sec:supp2}.

\begin{table}[htbp]
\centering
\footnotesize\setlength{\tabcolsep}{2.75pt}
\caption{PCK@0.10 (\%) for Direct RoMa (D), Gaussian warp fusion (G; weight $0.5$, bandwidth $0.05$), and full RelGeo (R). Bold marks the best result within each backbone.}
\label{tab:supp6}
\begin{tabular}{@{}lrrrrrr@{}}
\toprule
 & \multicolumn{3}{c}{SD2.1} & \multicolumn{3}{c}{FLUX} \\
\cmidrule(lr){2-4}\cmidrule(lr){5-7}
Setting & D & G & R & D & G & R \\
\midrule
SPair & 50.38 & 54.90 & \textbf{61.55} & 49.62 & 54.72 & \textbf{69.43} \\
CC & 68.00 & \textbf{68.87} & 68.02 & 68.00 & 70.10 & \textbf{76.38} \\
SOC & 62.58 & \textbf{63.11} & 58.02 & 62.58 & 63.85 & \textbf{64.52} \\
Cross-SOC & 40.23 & 42.01 & \textbf{48.82} & 40.23 & 42.78 & \textbf{56.32} \\
AP & 53.20 & \textbf{54.09} & 51.42 & 53.20 & 54.91 & \textbf{56.46} \\
\bottomrule
\end{tabular}
\end{table}

The forward transport prior is particularly competitive on SD2.1-SOC: Direct RoMa reaches 62.58\% and Gaussian fusion reaches 63.11\%, compared with 58.02\% for RelGeo. Gaussian fusion also leads on SD2.1-CC and AP-10K. RelGeo leads in the remaining seven comparisons, including all five FLUX settings and SD2.1 SPair/Cross-SOC. The relative strength of geometric endpoints and relationally calibrated similarity thus varies across backbones and benchmarks. WIC uses bidirectional endpoint compatibility and reliability to contribute geometric evidence to the same full-domain cosine decision as SRC and the frozen descriptors (main-paper Sec.~4.3).

\subsection{Random and Mismatched References}
The pair-alignment controls test whether relational coordinates convey information specific to the evaluated image pair. With Native descriptors and readout fixed, dimension-matched random SRC remains near Native, while same-class donor relations reduce accuracy (main-paper Table~\ref{tab:4}). Random SRC uses Gaussian coordinates seeded by image identifier, with the same dimension, per-location normalization, and $\lambda=0.35$. The mismatch condition substitutes another ordered pair's bilateral coordinate tensors without changing the current Native descriptors or readout.

The full-stack AP control keeps Native descriptors and SRC from the current pair and replaces the complete WIC reference. Adjacent image pairs within each class exchange tensors; the final unpaired entry uses the first pair, excluding itself. The swap transfers both endpoint encoding and donor reliability amplitudes, including their contribution to normalization.

\begin{table}[htbp]
\centering
\caption{Full RelGeo with current or mismatched WIC on AP-10K (\%). Both rows cover all 50 classes and 20,439 pairs.}
\label{tab:supp8}
\footnotesize
\resizebox{\columnwidth}{!}{%
\begin{tabular}{@{}lrrr@{}}
\toprule
Backbone & Current WIC & Mismatched WIC & Drop (points) \\
\midrule
SD2.1 & 51.42 & 44.18 & 7.24 \\
FLUX & 56.46 & 50.19 & 6.27 \\
\bottomrule
\end{tabular}
}
\end{table}

Replacing current-pair WIC with same-category references reduces full RelGeo accuracy for both backbones (Table~\ref{tab:supp8}).

\subsection{Success Sets of the Two Calibrated Branches}
\begin{figure}[tbp]
\centering
\includegraphics[width=\columnwidth]{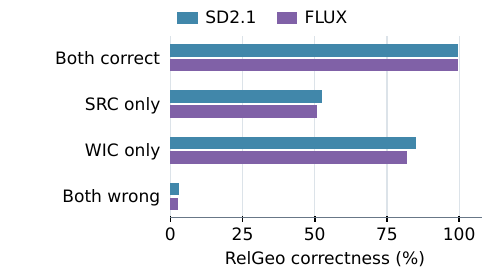}
\caption{RelGeo correctness across four SRC/WIC branch-success groups. Blue denotes DIFT-SD2.1 and purple denotes DiTF-FLUX. Bar lengths are pooled conditional correctness rates; group support counts and their interpretation are described in the text.}
\label{fig:supp_joint}
\end{figure}

The calibrated branches have overlapping but distinct success sets. We partition all 149,646 SOC queries by Native+SRC and Native+WIC correctness; Fig.~\ref{fig:supp_joint} reports full RelGeo accuracy and pooled support in each disjoint group. Group accuracies use pooled queries, while the main PCK results use class-macro aggregation.

Joint calibration retains many successes unique to either branch, while shared failures remain difficult. The WIC branch has 2.2--2.6 times as many uniquely correct queries as SRC. Full RelGeo retains 81.76--84.91\% of WIC-only successes and 50.63--52.63\% of SRC-only successes. Retention exceeds 99.5\% when both branches are correct; recovery stays below 3\% when both fail.

\section{Correspondence Identity Analysis}
\label{sec:supp7}

\begin{sloppypar}
RelGeo improves annotated identity for both backbones, with different contributions from SRC and WIC. RoleAcc evaluates all 108,875 eligible SOC queries in 100 classes with at least two visible target annotations for the source concept. The semantic-pair record specifies one correct identity; all others are distractors, including annotations inside the same PCK tolerance region. Responses are bilinearly sampled at normalized annotation locations with \texttt{align\_corners=}\allowbreak\texttt{True}. For positive response $s^+$ and distractor responses $s_j$, identity is correct if $s^+>\max_j s_j$, and rank is $1+\sum_j\mathbb I[s_j\ge s^+]$. Ties count against the positive. This discrete diagnostic differs from official SOC-geo and full-domain PCK.
\end{sloppypar}

\begin{table}[htbp]
\centering
\caption{SOC identity discrimination. RoleAcc, identity repair, and break are class-macro percentages; rank is lower-is-better. Repair/break condition on Native's incorrect/correct annotated identity.}
\label{tab:supp9}
\resizebox{\columnwidth}{!}{\begin{tabular}{@{}llrrrr@{}}
\toprule
Backbone & Branch & RoleAcc & Mean rank & Repair & Break \\
\midrule
SD2.1 & Native & 63.17 & 1.4908 & --- & --- \\
 & +SRC & 63.12 & 1.4853 & 6.30 & 3.77 \\
 & +WIC & 66.84 & 1.4378 & 18.18 & 4.04 \\
 & RelGeo & 66.91 & 1.4323 & 20.42 & 5.41 \\
FLUX & Native & 59.66 & 1.5135 & --- & --- \\
 & +SRC & 61.44 & 1.4896 & 10.42 & 3.67 \\
 & +WIC & 65.58 & 1.4342 & 23.24 & 4.88 \\
 & RelGeo & 66.29 & 1.4242 & 26.29 & 5.69 \\
\bottomrule
\end{tabular}}
\end{table}

WIC improves RoleAcc for both backbones, and adding SRC yields a further 0.07/0.71 points. SRC alone improves FLUX but stays close to Native for SD2.1: its identity effects vary with the representation despite positive PCK gains in the main ablation.

\section{Published Comparisons and Implementation Scope}
\label{sec:supp8}

\subsection{Published Comparisons and Appearance Fusion}
Main-paper Table~\ref{tab:published_spair_full} retains all 18 SPair classes. Its supervised 74.6\% row is the trained \emph{Fuse-ViT-B/14} model from A Tale of Two Features~\cite{ref5}, commonly denoted SD+DINO (S) in later comparisons, with manual correspondence supervision and source-reported per-image aggregation. SCOT~\cite{ref13} freezes model weights but uses validation keypoints for layer and parameter selection; its published score also uses per-image aggregation.

GeoAware-SC~\cite{ref6} retains pose alignment, index-assisted flipping, and windowed readout; its SD/DINO configurations use panoptic-segmentation-adapted CLIP conditioning. Fuse2Match~\cite{ref48} combines SD3.5-large, SD2.1, and DINOv2-large; its published Overall 68.8 is not the arithmetic mean of displayed class scores. CleanDIFT's TLR appearance setting reports bbox-normalized per-point PCK 69.99~\cite{ref3}; image-normalized 78.40 uses a different metric. AP-10K results in this paper use the intra-species test set and class-macro aggregation defined in Sec.~\ref{sec:supp1}.

Our independent appearance branch uses zero-based block 11 of frozen DINOv2-B/14. Features $E_i$ are bilinearly mapped to the $A_i$ grid with \texttt{align\_corners=False}; each component is normalized before concatenation:
\begin{equation}
B_i=\mathcal N([\mathcal N(A_i),\mathcal N(E_i)]),\qquad
R_i^{\rm DINO}=\mathcal N([B_i,I_i^{\rm WIC}]).
\label{eq:supp6}
\end{equation}
WIC retains its reliability amplitude. Appearance fusion follows variant-specific feature processing: Native+DINO jointly reduces FLUX diffusion features to 1,280 dimensions by PCA, whereas SRC-based variants use the hierarchical fusion above. The comparison therefore includes both relational coordinates and pipeline-specific feature processing.

\begin{table}[htbp]
\centering
\caption{Independent DINO appearance fusion on SPair-71k. Class-macro per-point PCK@0.10 (\%); both backbones use the corresponding native evaluation protocols.}
\label{tab:supp10}
\small
\setlength{\tabcolsep}{4pt}
\begin{tabular}{@{}llrr@{}}
\toprule
Setting & Branch & SD2.1 & FLUX \\
\midrule
SPair & Native+DINO & 65.95 & 67.22 \\
 & Native+SRC+DINO & 65.86 & 68.25 \\
 & RelGeo+DINO & 67.8 & 70.22 \\
\bottomrule
\end{tabular}
\end{table}

On SPair-71k, RelGeo+DINO exceeds Native+DINO by 1.85 points for SD2.1 and 3.00 points for FLUX. Adding SRC alone to Native+DINO changes PCK by $-0.09$ and $+1.03$ points, respectively.

\subsection{Computation and Reuse}
We profile three first-ordered SOCO-SOC pairs (aeroplane, ambulance, and armchair) at $800\times800$ on an NVIDIA A800 80GB PCIe, with RoMa fields cached. SD2.1 uses $E=1,r=128$ and FLUX uses $E=8,r=96$, both with \texttt{r64\_i2}. After warm-up, feature production is timed once per pair and calibration with cosine NN five times, using CUDA-synchronized wall-clock timing. Q/K is captured in the SRC/RelGeo forward; the profile includes GPU image transfer but excludes RoMa inference, image decoding, disk I/O, and model loading. Warp/support fields are cached on GPU, while endpoint encoding, cycle weighting, fusion, and readout are recomputed.

In this cached-field profile, the RelGeo stage sum increases by 5.03\%/8.01\% over Native for SD2.1/FLUX. RoMa fields can be stored and reused for repeated evaluations of the same pair; SPair uses different resolution and sampling configurations.

\begin{table}[!htbp]
\centering
\caption{Three-pair warm-model profile with cached RoMa. Times are ms/pair; stage sums add separately measured production and calibration. Peak is maximum allocated calibration memory in MiB.}
\label{tab:supp_runtime}
\resizebox{\columnwidth}{!}{\begin{tabular}{@{}llrrrr@{}}
\toprule
Backbone & Branch & Calib.+NN & Stage sum & Increase & Peak \\
\midrule
SD2.1 & Native & 1.10 & 325.6 & --- & 4869.5 \\
 & +SRC & 15.29 & 339.8 & 4.35\% & 4903.7 \\
 & +WIC & 3.08 & 327.6 & 0.61\% & 4885.4 \\
 & RelGeo & 17.48 & 342.0 & 5.03\% & 4919.5 \\
FLUX & Native & 1.10 & 3477.8 & --- & 33181.4 \\
 & +SRC & 145.95 & 3753.6 & 7.93\% & 33655.0 \\
 & +WIC & 4.62 & 3481.3 & 0.10\% & 33231.5 \\
 & RelGeo & 148.75 & 3756.4 & 8.01\% & 33655.0 \\
\bottomrule
\end{tabular}}
\end{table}

\section{Additional Qualitative Correspondence Results}
\label{sec:supp9}

\begin{figure*}[!p]
\centering
\includegraphics[width=.99\textwidth,height=.88\textheight,keepaspectratio]{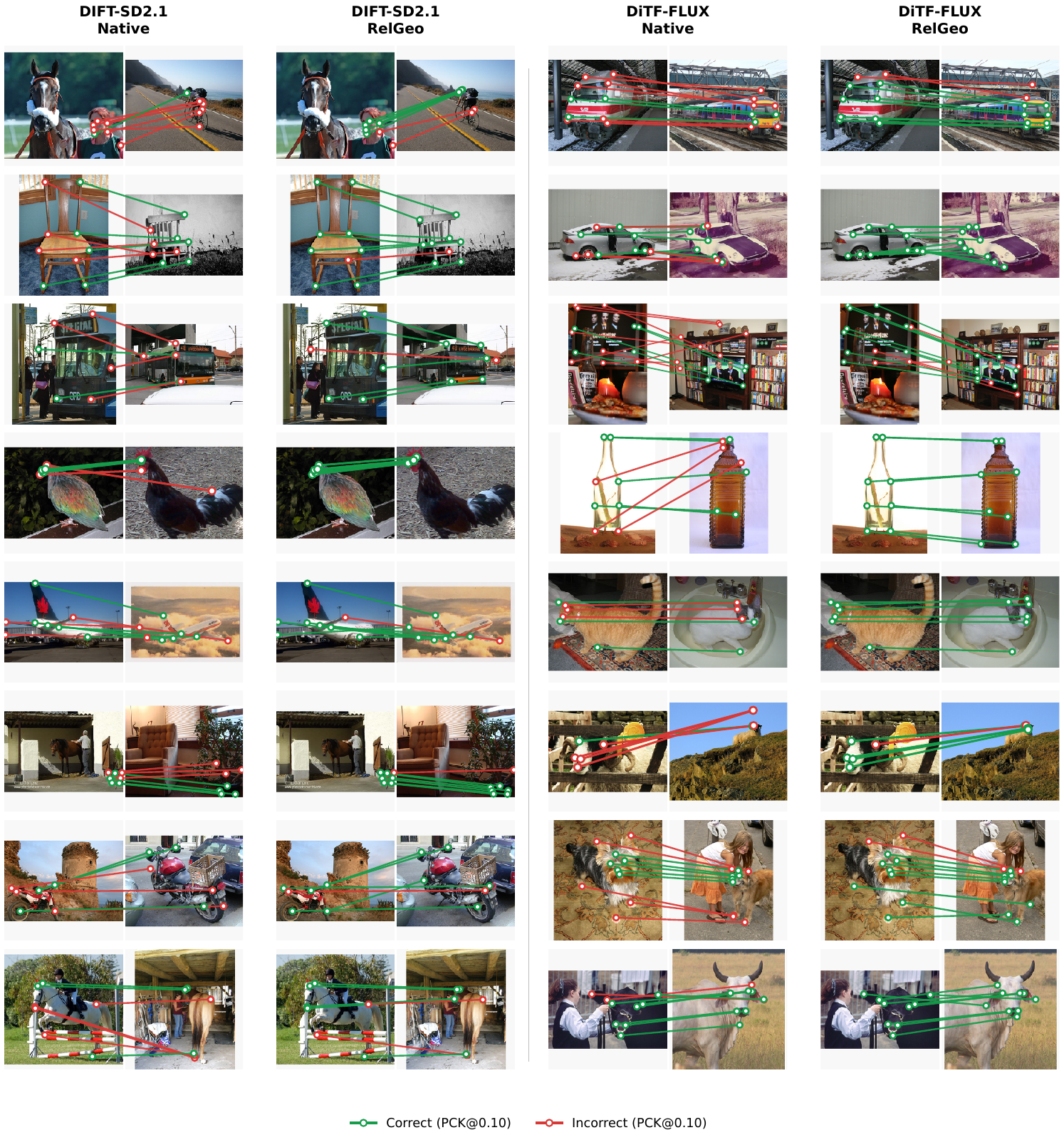}
\caption{Additional SPair-71k correspondences. Each eight-row backbone block compares Native and RelGeo for DIFT-SD2.1 (left) and DiTF-FLUX (right). Green/red indicate correct/incorrect PCK@0.10 matches.}
\label{fig:s9_spair}
\end{figure*}

\begin{figure*}[!p]
\centering
\includegraphics[width=.99\textwidth,height=.88\textheight,keepaspectratio]{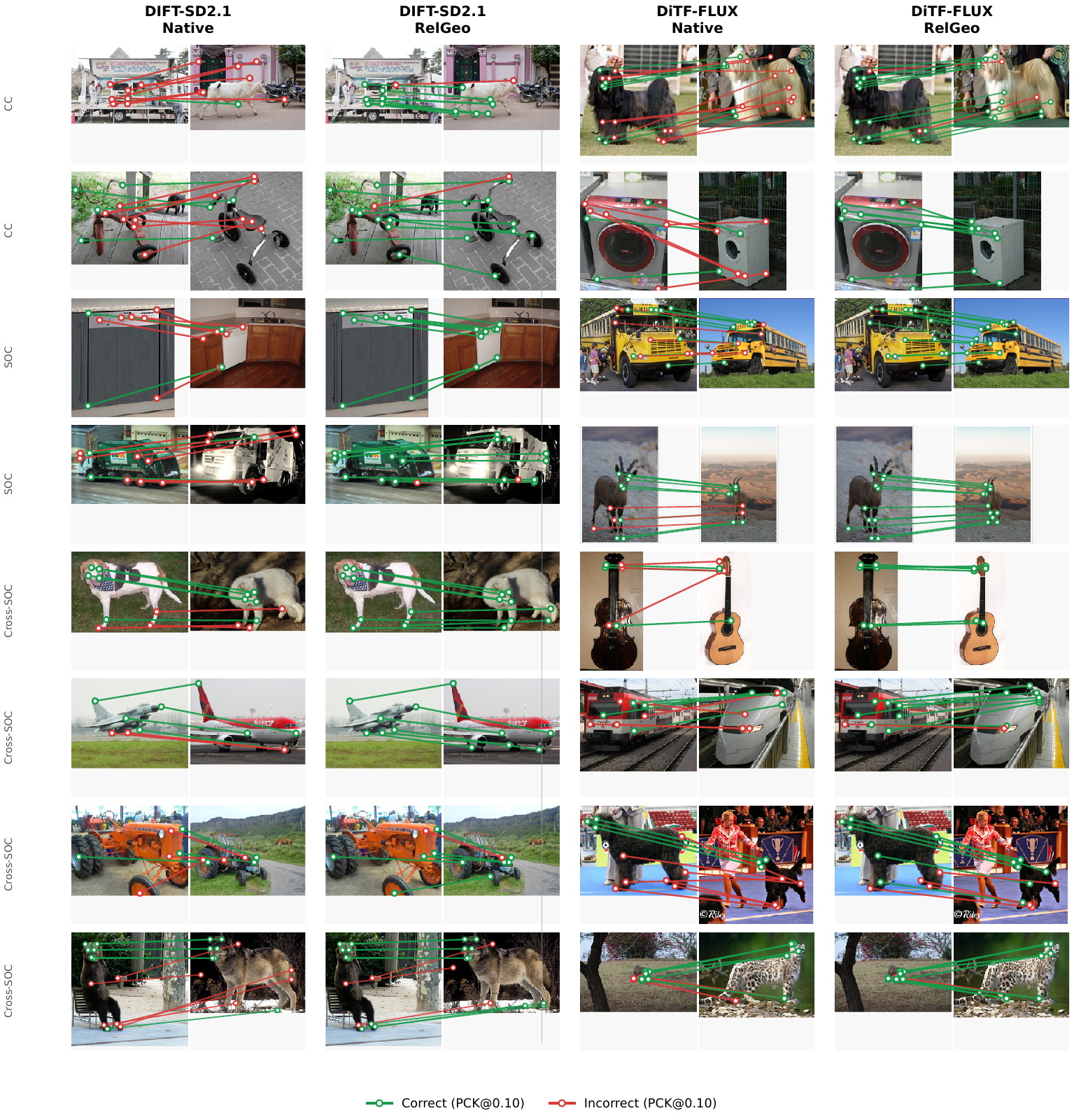}
\caption{Additional SOCO correspondence results for concept-level CC, same-object-relative-identity SOC, and Cross-SOC. Native and RelGeo are compared separately for DIFT-SD2.1 and DiTF-FLUX. Green/red indicate correct/incorrect PCK@0.10 matches.}
\label{fig:s9_soco}
\end{figure*}

\begin{figure*}[!p]
\centering
\includegraphics[width=.99\textwidth,height=.88\textheight,keepaspectratio]{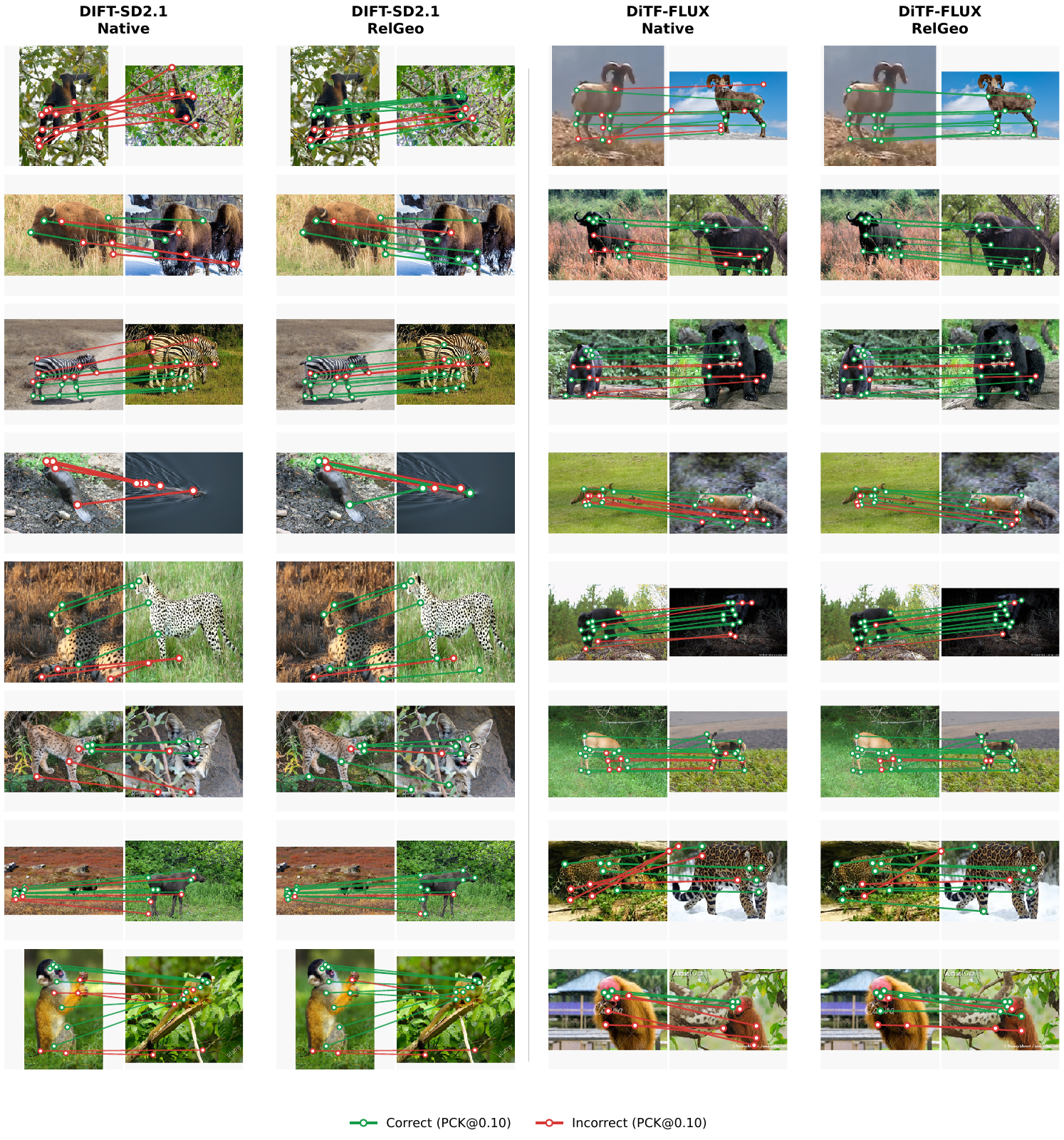}
\caption{Additional AP-10K intraspecies correspondences under articulated pose and appearance differences. Each backbone is compared using identical source queries for Native and RelGeo; green/red mark correct/incorrect matches under AP-10K PCK@0.10.}
\label{fig:s9_ap10k}
\end{figure*}

Figures~\ref{fig:s9_spair}--\ref{fig:s9_ap10k} extend main-paper Figure~\ref{fig:qualitative} across SPair-71k, SOCO, and AP-10K. Within each row, Native and RelGeo share the same images and queries. Green/red lines denote correct/incorrect matches under the corresponding PCK@0.10 criterion. These are diagnostic examples; quantitative conclusions use the full benchmarks.

\begin{figure*}[!p]
\centering
\includegraphics[width=.99\textwidth,height=.88\textheight,keepaspectratio]{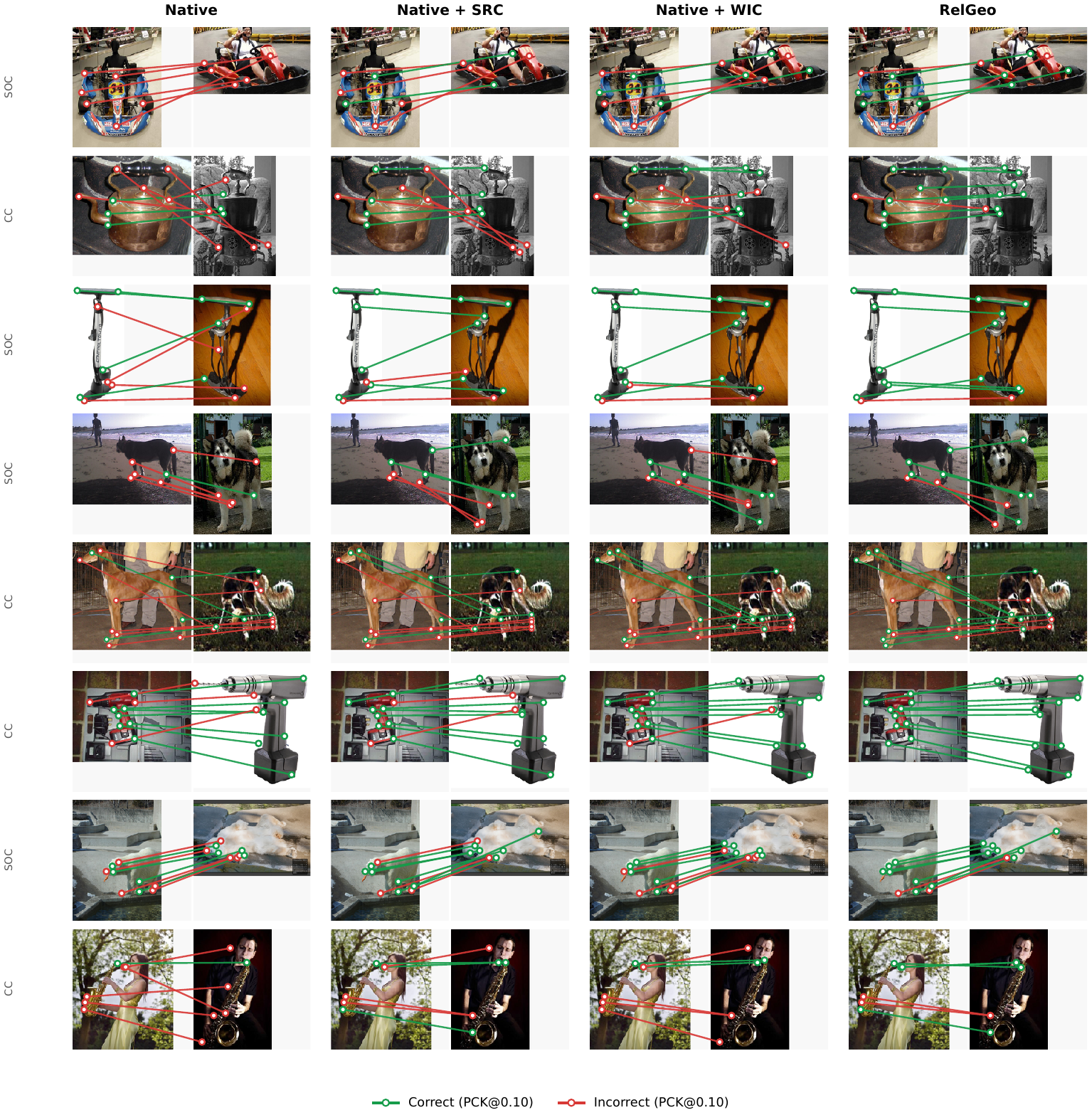}
\caption{DIFT-SD2.1 qualitative component ablation on SOCO. Columns show Native, Native+SRC, Native+WIC, and full RelGeo for the same queries. The comparisons illustrate the complementary contribution of the two pair-conditioned relation sources.}
\label{fig:s9_sd21_ablation}
\end{figure*}

\begin{figure*}[!p]
\centering
\includegraphics[width=.99\textwidth,height=.88\textheight,keepaspectratio]{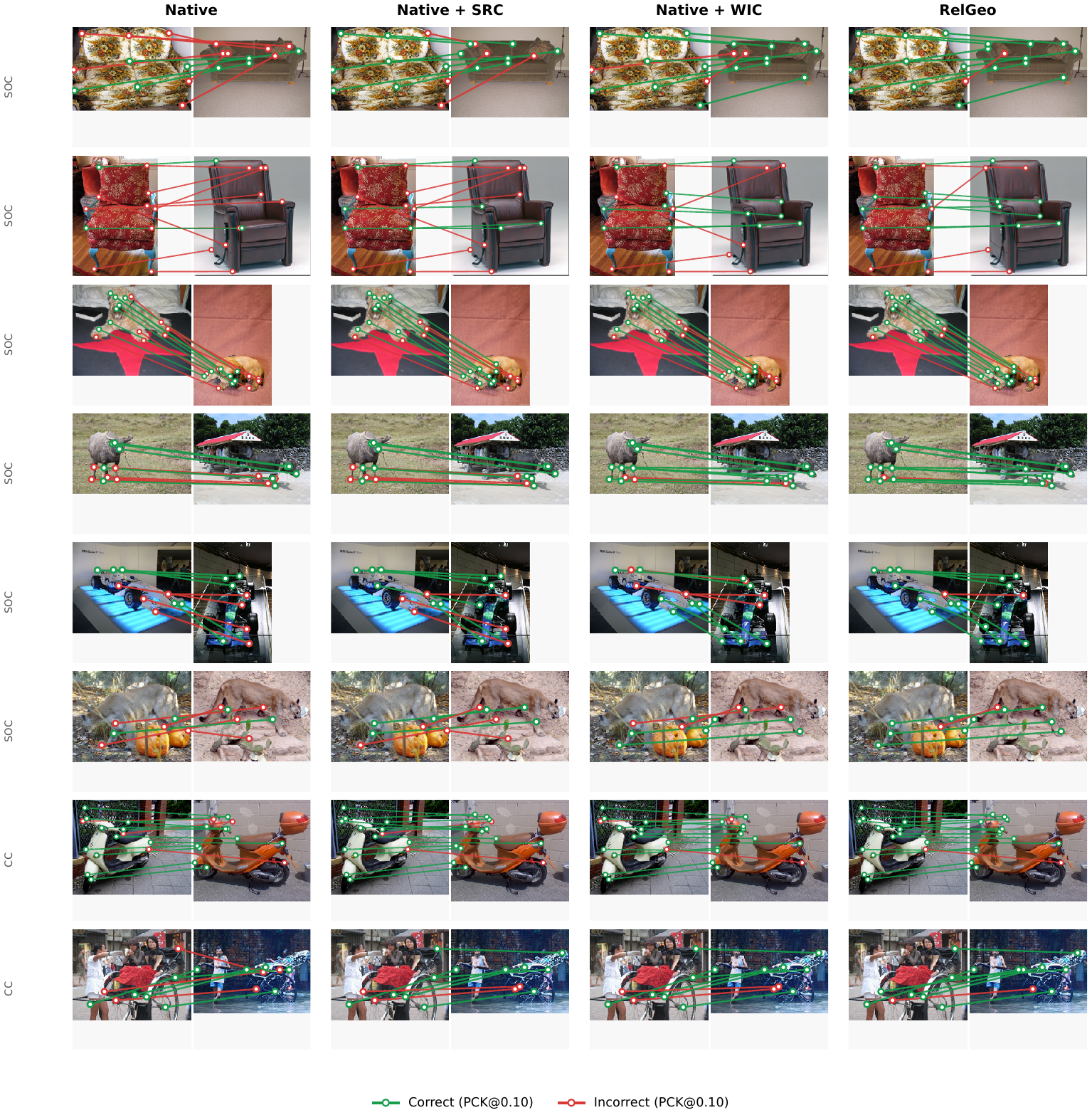}
\caption{DiTF-FLUX qualitative component ablation on SOCO. Columns show Native, Native+SRC, Native+WIC, and RelGeo under identical source queries and PCK@0.10 criteria.}
\label{fig:s9_flux_ablation}
\end{figure*}

Figures~\ref{fig:s9_sd21_ablation} and~\ref{fig:s9_flux_ablation} compare Native, +SRC, +WIC, and RelGeo for identical SOCO queries. Image-pair identities are distinct, with category repetition minimized across figures.

\clearpage

\end{document}